%% file: main.tex
\documentclass{article}

\usepackage[final, nonatbib]{neurips_2019}

\usepackage[utf8]{inputenc} %
\usepackage[T1]{fontenc}    %
\usepackage{hyperref}       %
\usepackage{url}            %
\usepackage{booktabs}       %
\usepackage{amsfonts}       %
\usepackage{nicefrac}       %
\usepackage{microtype}      %
\usepackage[dvipsname,svgnames,x11names,hyperref]{xcolor}

\hypersetup{
  colorlinks   = true, %
  urlcolor     = NavyBlue!50!Black, %
  linkcolor    = OrangeRed!85!Black, %
  citecolor   = Maroon %
}

\usepackage[style=phys, backend=biber, articletitle=false,biblabel=brackets,chaptertitle=false,pageranges=false]{biblatex}

\input{packages}

\input{definitions}

\title{Embedded-physics machine learning for coarse-graining and collective variable discovery without data}

\author{%
  Markus Schöberl\thanks{\url{https://mjschoeberl.github.io/},~ \url{http://orcid.org/0000-0001-7559-2619}} \\
  Technical University of Munich \\
  University of Notre Dame\\
  \texttt{mschoeberl@gmail.com} \\
  \AND
  Nicholas Zabaras\thanks{311 Cushing Hall, Notre Dame, IN 46556, USA. \url{http://orcid.org/0000-0003-3144-8388}} \\
  Center for Informatics and Computational Science\\
  University of Notre Dame \\
  \texttt{nzabaras@gmail.com} \\
  \And
  Phaedon-Stelios Koutsourelakis\thanks{Boltzmannstra{\ss}e 15, 85748 Garching, Germany. \url{http://orcid.org/0000-0002-9345-759X}} \\
  Professorship of Continuum Mechanics \\
  Technical University of Munich \\
  \texttt{p.s.koutsourelakis@tum.de} \\
}

\begin{document}

\maketitle

\begin{abstract}
We present a novel learning framework that consistently embeds underlying physics while bypassing a significant drawback of most modern, data-driven coarse-grained approaches in the context of molecular dynamics (MD), i.e., the availability of big data. The generation of a sufficiently large training dataset poses a computationally demanding task, while complete coverage of the atomistic configuration space is not guaranteed. As a result, the explorative capabilities of data-driven coarse-grained models are limited and may yield biased ``predictive'' tools. We propose a novel objective based on reverse Kullback--Leibler divergence that fully incorporates the available physics in the form of the atomistic force field. Rather than separating model learning from the data-generation procedure - the latter relies on simulating atomistic motions governed by force fields - we query the atomistic force field at sample configurations proposed by the predictive coarse-grained model. Thus, learning relies on the evaluation of the force field but does not require any MD simulation. The resulting generative coarse-grained model serves as an efficient surrogate model for predicting atomistic configurations and estimating relevant observables. Beyond obtaining a predictive coarse-grained model, we demonstrate that in the discovered lower-dimensional representation, the collective variables (CVs) are related to physicochemical properties, which are essential for gaining understanding of unexplored complex systems. We demonstrate the algorithmic advances in terms of predictive ability and the physical meaning of the revealed CVs for a bimodal potential energy function and the alanine dipeptide.
\end{abstract}

\input{chapter_6}

\clearpage

\appendix

\input{appendix}

\medskip

\printbibliography[heading=bibintoc]

\end{document}

%% file: packages.tex
\usepackage{amsmath, amssymb}
\usepackage{amsfonts}
\usepackage{bm}
\usepackage{siunitx}
\usepackage{pdfpages} 
\usepackage{enumitem}
\usepackage{verbatim}

\usepackage{caption}
\usepackage{subfigure}

\usepackage[section]{placeins}

\usepackage{bibentry}

\usepackage{pgfplots}
\usepackage{transparent}

\usepackage{tikz}
\usepackage{tkz-tab}
\usetikzlibrary{bayesnet}
\usetikzlibrary{shapes,arrows,
positioning, %
calc %
}
\usetikzlibrary{backgrounds}

\usepackage{hyphenat}


\usepackage[linesnumbered,ruled,vlined]{algorithm2e}%
\DontPrintSemicolon
\SetKwRepeat{Do}{do}{while}
\let\oldnl\nl%
\newcommand{\nonl}{\renewcommand{\nl}{\let\nl\oldnl}}%

%% file: definitions.tex
\newcommand\be{\begin{equation}}
\newcommand\ee{\end{equation}}

\definecolor{mDarkBrown}{HTML}{604c38}
\definecolor{mDarkTeal}{HTML}{23373b}
\definecolor{mLightBrown}{HTML}{EB811B}
\definecolor{mLightGreen}{HTML}{14B03D}

\newcommand{\bx}{\mathbf{x}}
\newcommand{\bz}{\mathbf{z}}

\newcommand{\btheta}{\boldsymbol{\theta}}
\newcommand{\bphi}{\boldsymbol{\phi}}

\newcommand{\bmu}{\boldsymbol{\mu}}

\newcommand\bsig{\boldsymbol{\sigma}}
\newcommand\beps{\boldsymbol{\epsilon}}

\newcommand{\data}{{\mathcal{D}_N}}

\newcommand{\indf}{_{\text{f}}}
\newcommand{\indcc}{_{\text{c}}}
\newcommand{\indcf}{_{\text{cf}}}
\newcommand{\indtarg}{_{\text{target}}}
\newcommand{\indthetacf}{_{\btheta\indcf}}
\newcommand{\indthetacc}{_{\btheta\indcc}}
\newcommand{\indtheta}{_{\btheta}}
\newcommand{\indphi}{_{\bphi}}
\newcommand{\bzi}{\boldsymbol{z}^{(i)}}
\newcommand{\bxi}{\boldsymbol{x}^{(i)}}

\newcommand{\ibrack}{{(i)}}
\newcommand{\eqqref}[1]{Equation \ref{#1}}
\newcommand{\secref}[1]{Section \ref{#1}}
\newcommand{\figref}[1]{Figure \ref{#1}}
\newcommand{\hleq}[1]{\colorbox{mDarkTeal!10}{$\displaystyle#1$}}

%% file: chapter_6.tex
Boltzmann densities, which are ensemble representations of equilibrium atomistic systems, are usually explored by molecular dynamics (MD) \cite{adler1959_md} or Monte Carlo-based (MC) techniques \cite{metropolis1949_mc}. These versatile and general simulation techniques asymptotically guarantee unbiased estimates of observables \cite{weiner2012_statmech}. However, these simulation techniques become computationally impractical in cases where the atomistic interaction potential exhibits several distinct minima or wells. Such complex potentials imply multimodal Boltzmann densities. Escaping such a well is rare and requires overcoming high free-energy barriers, resulting in impractically long simulation times or biased trajectories \cite{maisuradze2010_free_energy_rugged}. 

Key to exploring such multimodal Boltzmann densities is the recognition of appropriate slow coordinates or collective variables (CVs) that exhibit sensitivity in transition regions between modes. This requires tremendous physicochemical insight, which is not available per se. CVs, which provide an effective lower-di\-men\-sio\-nal description of high-dimensional atomistic systems, are key to accelerating the exploration of multimodal densities by biasing the dynamics to escape deep free-energy wells \cite{rohrdanzclementi2013}.

Identifying expressive CVs governing major conformational changes in the absence of physical insight requires data-driven strategies. However, in many cases, the identification of CVs requires a dense sample of the target Boltzmann distribution, as well as unbiased simulation trajectories. This creates a contradiction, as it is computationally impractical to obtain unbiased trajectories in the presence of multiple modes \cite{sittel2018_cv_nonlinear_map}, restricting the potential use-cases of such approaches. 
Few nonlinear dimensionality reduction methods coping with low data provide CVs that are not differentiable with respect to their atomistic counterparts \cite{balasubramanian2002, donoho2003, risken1996}; however, this is required for biasing the dynamics  \cite{tenenbaum2000, ceriotti2011, rohrdanz2011, coifman2005, coifman2005b, ferguson2011, nadler2006, nadler2008, noe2013, noe2016, mccarty2017, zheng2013}.
Deep learning approaches providing flexible and efficiently differentiable functions have also influenced research on the efficient exploration of multimodal Boltzmann distributions. However, these build on previously acquired reference data and do not account directly for the interaction potential that actually drives the MD or MC simulation.

This work provides a novel and fundamentally different perspective on data-driven deep learning approaches. 
Instead of relying on two separate process, acquiring data and then employing statistical learning of a model, we synthesize and embed physics, i.e., the Boltzmann density, with a machine learning objective.
The advocated learning methodology proposes (atomistic) configurations to which the model is attracted to learn from the potential energy and associated interatomic forces. The proposed machine learning algorithm does not require any simulation of the Boltzmann density but only queries the physical model, i.e., the potential and forces, to gain relevant information by evaluating rather than simulating the Boltzmann density.
The proposed learning algorithm includes a versatile nonlinear dimensionality-reduction routine, which simultaneously discovers relevant CVs while learning the Boltzmann density. We demonstrate the procedure using a double well potential and the alanine dipeptide.

The present work differs clearly from recent developments on Boltzmann generators \cite{noe2019_boltzmann_gen} that rely on invertible neural networks such as RealNVP \cite{dinh2017_realnvp} and NICE \cite{dinh2015_nice}. As it employs invertible neural networks, the dimensionality of the latent generator must equal the dimensionality of the atomistic configuration, which detains a consistent dimensionality reduction. 
Generated atomistic realizations of the employed model in  \cite{noe2019_boltzmann_gen} do not reflect the statistics of the reference Boltzmann distribution and serve instead as an input to a subsequent re-weighting importance sampling step. However, importance sampling is difficult to monitor if the variance in the importance weights is low, implying a large effective sample size, when none of the proposed realizations yield relatively high probabilities as evaluated by the target density \cite{liu2008}. Furthermore, a good guess of CVs is provided in Boltzmann generators, which depict physical insights that may not be available.
By contrast, the proposed approach (similar double well example) reveals the effective CVs \emph{and} also provides a generator to produce samples that yield the correct statistics of the target.

In the following \secref{sec:ch5_var_cg_methodology}, we develop the proposed learning approach based on KL divergence minimization and derive a tractable upper bound based on hierarchical variational models \cite{ranganath2016_hvm}. We discuss the required gradient computation and provide a physically interpretable underpinning of the components involved. After introducing a general model parametrization, we provide an adaptive tempering scheme facilitating a robust machine learning procedure at the end of \secref{sec:ch5_var_cg_methodology}.  The proposed physics-embedding learning procedure for revealing CVs and obtaining a coarse-grained (CG) model is numerically validated in \secref{sec:ch5_nummerical_illustrations} with a double well potential and the alanine dipeptide. We close this paper in \secref{sec:ch5_conclusions}, summarizing the main findings of this work and outlining simple but effective further extensions and research directions. These include the generalization of the obtained predictive distribution for predictive purposes at any temperature.

\section{Methodology}
\label{sec:ch5_var_cg_methodology}

After introducing the notation in \secref{sec:ch5_methods_stat_mech}, we describe the general proposed framework in \secref{sec:ch5_methods_pgm}. A tractable optimization objective is provided in \secref{sec:ch5_methods_inference}. We compare the proposed approach with data-driven objectives in \secref{sec:ch5_methods_comparison_var_data}. Relevant model specifications and gradient computations for training are discussed in \secref{sec:ch5_methods_model_gradient}, and we close with some notes on the actual training procedure in \secref{sec:ch5_methods_training}.

\subsection{Equilibrium statistical mechanics}
\label{sec:ch5_methods_stat_mech}

In equilibrium statistical mechanics, we seek to estimate ensemble averages of observables $a(\bx)$ with respect to the Boltzmann density,
\begin{equation}
    \left\langle a \right\rangle_{p\indtarg(\bx; \beta)} = \int_{\mathcal{M}\indf} a(\bx) p\indtarg(\bx;\beta) ~d\bx
    \label{eqn:ch5_phase_average}.
\end{equation}
We denote the Boltzmann distribution, for which we aim to learn an efficient approximation, by $p\indtarg(\bx;\beta)$:
\begin{align}
        p\indtarg(\bx) &= \frac{1}{\mathrm{Z}(\beta)} \underbrace{e^{-\beta U(\bx)}}_{\pi(\bx;\beta)}
        \label{eqn:ch5_methods_boltzmann} \\ \nonumber
    &= \frac{\pi(\bx;\beta)}{\mathrm{Z}(\beta)}.
\end{align}
In \eqqref{eqn:ch5_methods_boltzmann},  $\mathrm{Z}(\beta) = \int e^{-\beta U(\bx)} d\bx$ is the partition function or normalization constant, and $\beta = \frac{1}{k_B T}$ is  the reciprocal or inverse temperature with the Boltzmann constant $k_B$ and the temperature $T$. The interatomic potential $U(\bx)$ depends on generalized atomistic  coordinates denoted by $\bx \in \mathcal M_\text{f} \subset \mathbb R^{n\indf}$, with $n\indf=\dim(\bx)$. 
In equilibrium statistical mechanics, we are usually interested in phase averages at distinct constant temperatures; however, we will also demonstrate how to  utilize the temperature to introduce an auxiliary sequence of target distributions to facilitate learning the actual target distribution. The auxiliary sequence stabilizes the parameter learning inspired by annealing \cite{lecchini2008, li2009-ow} and adaptive sequential MC \cite{bilionis2012_free_energy}.

\subsection{Coarse-graining through probabilistic generative models}
\label{sec:ch5_methods_pgm}

Data-driven coarse-graining methodologies are based on a limited set of $N$ realizations obtained from the target density $p\indtarg(\bx)$. The realizations $\bxi$ are produced by drawing samples from  $p\indtarg(\bx)$: $\bxi \sim p\indtarg(\bx)$ with Markov Chain MC (MCMC) methods \cite{metropolis1953, grenander1994_langevin} and/or, especially in the context of biochemical atomistic systems, by MD simulations \cite{allen2017_md_basics, frenkel1996}. Both methodologies yield a dataset $\bx^\data = \{\bxi\}^N_{i=1}$, which approximates the target distribution with
\begin{align}
    p\indtarg(\bx) &\approx \Tilde{p}(\bx) \nonumber \\
    &\propto \prod_{i=1}^N \delta(\bxi - \bx).
    \label{eqn:ch5_methods_approx_data}
\end{align}
The above approximation, given independent and identically distributed samples, may sufficiently resemble simple systems. However, atomistic many-body systems exhibit higher-order and long-range interactions \cite{tuckerman2000_md_overview, shaw2014} involving multiple free energy modes separated by high barriers \cite{bernardi2015_enhanced, karplus1979_protein_folding}. Therefore, the collection of sufficient data becomes an insurmountable task: a protein folding process may take microseconds versus a time discretization of femtoseconds \cite{freddolino2010_folding_challenges}. Given limited computational power, the relevant conformations and transitions are not guaranteed to be reflected by the reference simulation \cite{ensign2009_sufficient_reference_trj}. 

The quality of data-driven learning approaches depends strongly on the quality of the available set of reference data $\bx^\data$. If, e.g., in the case of peptides, certain conformations are missed, it is an almost insurmountable challenge to obtain a data-driven model exploring such missed configurations \cite{chen2017_data_cg, laio2002_escaping_minima}. Enhanced sampling methods \cite{tenenbaum2000, ceriotti2011, rohrdanz2011, coifman2005, coifman2005b, ferguson2011, nadler2006, nadler2008, noe2013, noe2016, mccarty2017, zheng2013} can support the exploration of the configuration space, while the efficiency crucially depends on the quality of utilized CVs \cite{rohrdanz2013_cv_discovery, pande2017}.

Instead of relying on reference data, which may be a distorted representation of $p\indtarg(\bx)$, or gradually exploring the configuration space by enhanced sampling, we present a variational approach that learns the target distribution $p\indtarg(\bx)$ by querying the unnormalized distribution $\pi(\bx)$ or the corresponding potential energy $U(\bx)$ (see \eqqref{eqn:ch5_methods_boltzmann}).

We are first interested in identifying latent CVs $\bz$ ($\bz \in \mathcal{M}\indcc \subset \mathbb{R}^{n\indcc}$) depending on the fully atomistic picture $\bx$, which encode physically relevant characteristics (e.g., coordinates along transition regions between conformations) and provide insight into the unknown atomistic system we seek to explore. Second, we seek to identify a CG model expressed in terms of the latent CVs $\bz$ that is predictive but nevertheless facilitates reasoning about all-atom coordinates $\bx$ \cite{schoeberl2017_pcg}. The obtained CG model is expected to serve as an approximation to $p\indtarg(\bx)$ to enable the efficient computation of the expectations of observables (\eqqref{eqn:ch5_phase_average}) and most importantly to capture relevant configurations in the free energy landscape that were inaccessible by brute-force MD or MCMC approaches \cite{grubmueller1995_conf_flodding}. Fulfilling the latter requirement also implies capturing statistics of $p\indtarg(\bx)$.

The CVs $\bz$ serve as latent generators of the higher-dimensional generalized coordinates $\bx$, where we seek $\dim(\bz) \ll \dim(\bx)$. 
This generative process is expressed with two components,
\begin{enumerate}[label=(\roman*)]
    \item the conditional density $q\indthetacf(\bx |\bz)$, parametrized by $\btheta\indcf$,
    \item and the density over the latent CVs $q\indthetacc(\bz)$.
\end{enumerate}
Combining both  densities gives the following joint: %
\begin{equation}
    q\indtheta(\bx,\bz) = q\indthetacf(\bx|\bz) q\indthetacc(\bz).
\label{eqn:ch5_method_joint}
\end{equation}
Assuming we have obtained the optimal parameters $\btheta^{\text{opt}}$ after a training process based on an objective, which we will discuss later in this section, we can utilize the model for predictive purposes. This can be done by ancestral sampling \cite{liu2008}, i.e., first draw $\bzi \sim q_{\btheta^{\text{opt}}}(\bz)$ and second $\bxi \sim q_{\btheta^{\text{opt}}}(\bx|\bzi)$.

For obtaining optimal parameters $\btheta$, many methods rely on minimizing a distance from the target distribution $p\indtarg(\bx)$ to the marginal distribution $q\indtheta(\bx)$, which is given by:
\begin{equation}
    q\indtheta(\bx) = \int q\indtheta(\bx, \bz) ~d\bz = \int q\indthetacf(\bx|\bz) q\indthetacc(\bz)~d\bz.
    \label{eqn:ch5_method_marginal}
\end{equation}
A commonly employed metric expressing the deviation between two densities is the Kullback--Leibler (KL) divergence, which belongs to the family of  $\alpha$-divergences \cite{adrzej2010, cha2007, lobato2016}:
\begin{align}
    D_\text{KL}\left( p\indtarg(\bx) \| q\indtheta(\bx) \right) &= -\int p\indtarg(\bx) \log \frac{q\indtheta(\bx)}{p\indtarg(\bx)} ~d\bx \nonumber \\
    &= -\left\langle \log q\indtheta(\bx) \right\rangle_{p\indtarg(\bx)} + \underbrace{ \hleq{\left\langle  \log p\indtarg(\bx) \right\rangle_{p\indtarg(\bx)}}}_{-\mathbb{H}(p\indtarg)}.
    \label{eqn:ch5_method_forward_kl}
\end{align}
Minimizing \eqqref{eqn:ch5_method_forward_kl} with respect to $\btheta$ leads to $q\indtheta(\bx)$ being closer to $p\indtarg(\bx)$. However, in practice, the expectations in \eqqref{eqn:ch5_method_forward_kl} are intractable:
\begin{enumerate}[label=(\roman*)]
    \item the marginal $q\indtheta(\bx)$ requires the integration with respect to $\bz$ which is intractable itself and
    \item the involved expectation with respect to $p\indtarg(\bx)$, $\left\langle \cdot \right\rangle_{p\indtarg(\bx)}$ is analytically intractable since the normalization constant of $p\indtarg(\bx)$ is unavailable (which would require solving an integral with respect to $\bx$).
\end{enumerate}
Considering the above challenges, the latter could be addressed by approximating $p\indtarg(\bx)$ with data or samples $\bx^\data$ and thus approximating the corresponding expectations with MC estimators. However, as we deal with complex multimodal Boltzmann densities  $p\indtarg(\bx)$, the data generating process (MCMC  or MD) may miss relevant modes. By employing a biased set of samples or data not approximating $p\indtarg(\bx)$ \cite{koutsourelakis2016_uq_predictive}, we learn a biased estimator not approximating $p\indtarg(\bx)$. The generation of the training dataset is thus decoupled from the learning process.

To circumvent the data-generating process and thus sampling from $p\indtarg(\bx)$, we propose employing the other extreme of the family of  $\alpha$-divergences (as compared to \eqqref{eqn:ch5_method_forward_kl}), the \emph{reverse}  KL divergence:
\begin{align}
    D_\text{KL}\left( q\indtheta (\bx) \| p\indtarg (\bx) \right) &= - \int q\indtheta (\bx)\log \frac{p\indtarg (\bx) }{q\indtheta(\bx) } ~d\bx \nonumber \\
    &= \underbrace{\hleq{\mathbb{E}_{q\indtheta(\bx)}\left[ \log q\indtheta(\bx) \right]}}_{-\mathbb{H}(q(\bx))} -  \mathbb{E}_{q\indtheta(\bx)}\left[ \log p\indtarg(\bx) \right] \nonumber \\
    &=  -  \mathbb{E}_{q\indtheta(\bx)}\left[ \log p\indtarg(\bx) \right] -\mathbb{H}(q(\bx)).
    \label{eqn:ch5_method_reverse_kl}
\end{align}
Minimizing \eqqref{eqn:ch5_method_reverse_kl} with respect to $\btheta$ requires maximizing the log-likelihood $\log p\indtarg(\bx)$ assessed under $q\indtheta(\bx)$ (first component in \eqqref{eqn:ch5_method_reverse_kl}), and the maximization of the entropy of $q\indtheta(\bx)$, $\mathbb{H}(q(\bx))$ (second component in \eqqref{eqn:ch5_method_reverse_kl}). Minimizing the reverse KL divergence balances the two terms, as maximizing only the log-likelihood $\log p\indtarg(\bx)$ assessed under $q\indtheta(\bx)$ would result in a degenerate case where $q\indtheta(\bx)$ would become  a Dirac-delta placed at the (global) maximum of $p\indtarg(\bx)$ obtained at the (global) minimum of $U(\bx)$. The second component implies a regularization favoring  a parametrization $\btheta$ such that the entropy of $q\indtheta(\bx)$ is maximized.

\subsection{Inference and learning}
\label{sec:ch5_methods_inference}

In what follows, we use the negative of the KL divergence in \eqqref{eqn:ch5_method_reverse_kl} to be maximized, which we denote with $\mathcal{L}$ for the sake of comparability with other learning approaches \cite{rezendre2015_vi_normalizing_flow, kingma2014_autoencoding_vb}. At the end of this section, we compare the presented methodology with data-driven approaches relying on the forward KL divergence \cite{hoffman2013_stochastic_vi, ranganath2014_bbvi, kingma2014_autoencoding_vb} and especially those addressing coarse-graining problems \cite{hernandez2017_vae, noe2018_autoencoder_timelagged, mohamed2018_vae, schoeberl2019_pcvs}.

The objective to be maxmized is
\begin{equation}
    \mathcal L(\btheta) = \mathbb{E}_{q\indtheta(\bx)}\left[ \log p\indtarg(\bx) - \log q\indtheta(\bx) \right],
    \label{eqn:ch5_methods_elbo}
\end{equation}
where we can draw samples from $q\indtheta(\bx)$ as we can wisely select tractable hierarchical components composing to $q\indtheta(\bx)$. The optimization of the first component in $\mathcal{L}(\btheta)$ relating to the log-likelihood is tractable as the normalization of $p\indtarg(\bx) $ does not depend on the parameters $\btheta$ and thus being able to evaluate $\pi(\bx)$ or $U(\bx)$ suffices. However, the entropy term is not tractable ad-hoc as it involves the marginal  $q\indtheta(\bx) = \int q\indthetacf(\bx|\bz) q\indthetacc(\bz) ~d\bz$, posing in most cases an intractable or least cumbersome task.

Therefore,  we seek to construct a tractable lower bound on $\mathbb{H}(\bx)$ as presented in \cite{ranganath2016_hvm} by introducing an auxiliary density $r\indphi(\bz | \bx)$ parametrized by $\phi$ and write:
\begin{align}
    -\mathbb{E}_{ q(\bx)} \left[ \log  q(\bx) \right] &=  -\mathbb{E}_{ q(\bx)} \left[ \log  q(\bx) + \underbrace{D_\text{KL}\left( q(\bz|\bx) \| q(\bz|\bx) \right)}_{= 0} \right] \nonumber \\ 
    &\geq -\mathbb{E}_{ q(\bx)} \left[ \log  q(\bx) + D_\text{KL}\left( q(\bz|\bx) \|  r\indphi(\bz |\bx) \right) \right] \nonumber \\
    &= -\mathbb{E}_{ q(\bx)} \left[ \mathbb{E}_{q(\bz|\bx)} \left[ \log  q(\bx) + \log q(\bz|\bx) - \log r\indphi(\bz |\bx) \right] \right].
    \label{eqn:ch5_methods_entropy_lowerbound_1}
\end{align}
Adding $D_\text{KL}\left( q(\bz|\bx) \| q(\bz|\bx) \right)$ in the first line of  \eqqref{eqn:ch5_methods_entropy_lowerbound_1} has no influence as the term is equal to zero. It involves the posterior distribution over the latent variables $\bz$, $q(\bz|\bx) = \frac{q(\bx,\bz)}{q(\bx)}$, which is intractable.
By utilizing an auxiliary distribution $r\indphi(\bz | \bx)$, the equality becomes an inequality as a consequence of  $D_\text{KL}\left( q(\bz|\bx) \| r\indphi(\bz | \bx) \right) \geq 0$  for $r\indphi(\bz | \bx)$ deviating from $q(\bz|\bx)$.
Replacing the exact log-posterior $\log q(\bz|\bx)$ by
\begin{equation}
    \log q(\bz|\bx) = \log q(\bz) + \log q(\bx|\bz) - \log q(\bx),
\end{equation}
it follows that
\begin{align}
    -\mathbb{E}_{q(\bx)} \left[ \log  q(\bx) \right] \geq
    &-\mathbb{E}_{ q(\bx)} \Big[ \mathbb{E}_{q(\bz|\bx)} \big[ \hleq{\log  q(\bx)} + \log q(\bz) + \log q(\bx|\bz) \nonumber \\ &\hleq{ - \log  q(\bx) } - \log r\indphi(\bz |\bx)  \big] \Big] \nonumber \\
    = &-\mathbb{E}_{ q(\bx)} \left[ \mathbb{E}_{q(\bz|\bx)} \left[  \log q(\bz) + \log q(\bx|\bz) - \log r\indphi(\bz |\bx)  \right] \right].
    \label{eqn:ch5_methods_entropy_lowerbound}
\end{align}
Rewriting the expectation $\mathbb{E}_{ q(\bx)} \left[ \mathbb{E}_{q(\bz|\bx)} \left[ \cdot \right] \right]$ as $\mathbb{E}_{ q(\bx,\bz)} \left[ \cdot \right] $, 
\eqqref{eqn:ch5_methods_entropy_lowerbound} depicts a tractable lower bound on the entropy term. Maximizing the lower bound in \eqqref{eqn:ch5_methods_entropy_lowerbound} with respect to $\bphi$ minimizes $D_\text{KL}\left( q(\bz|\bx) \| r(\bz | \bx; \bphi) \right)$ and thus tightens the bound on the entropy term. As mentioned earlier, the optimum\footnote{The optimum with respect to $r\indphi(\bz |\bx)$ and thus $\bphi$ that tightens the lower bound.} is obtained when we identify the exact posterior of the latent CVs $q(\bz|\bx)$ with $r(\bz|\bx;\bphi^\text{opt}) = q(\bz|\bx)$, thus $D_\text{KL}\left( q(\bz|\bx) \| r(\bz|\bx;\bphi^\text{opt}) \right) = 0$.
Utilizing the obtained bound in \eqqref{eqn:ch5_methods_entropy_lowerbound}, the objective $\mathcal{L}(\bphi,\btheta)$ from \eqqref{eqn:ch5_methods_elbo} becomes:
\begin{equation}
    \mathcal L(\bphi, \btheta) = \mathbb{E}_{q(\bx,\bz;\btheta)}\left[ \log p\indtarg(\bx) - \log q\indthetacc(\bz) - \log q\indthetacf(\bx|\bz) + \log r\indphi(\bz |\bx)  \right].
    \label{eqn:ch5_methods_elbo_w_bound}
\end{equation}

The following shows the connection between the obtained objective and the KL divergence defined between the joint $q(\bx|\bz)q(\bz)$ and $p\indtarg(\bx)r(\bz|\bx)$ acting on the extended probability space:
\begin{align}
    \mathcal L(\bphi, \btheta) &= \mathbb{E}_{q(\bx,\bz;\btheta)}\left[ \log p\indtarg(\bx) - \log q(\bz) - \log q(\bx|\bz) + \log r\indphi(\bz |\bx)  \right] \nonumber \\
    &=\mathbb{E}_{q(\bx,\bz;\btheta)}\left[ \log \frac{ p\indtarg(\bx) r\indphi(\bz |\bx)}{q\indthetacf(\bx|\bz) q\indthetacc(\bz)}  \right] \label{eqn:ch5_methods_joint} \\
    &= - D_\text{KL}\left( q\indthetacf(\bx|\bz) q\indthetacc(\bz) \| p\indtarg(\bx) r\indphi(\bz |\bx) \right).
    \label{eqn:ch5_methods_KL_joint}
\end{align}
Based on \eqqref{eqn:ch5_methods_joint} and \eqqref{eqn:ch5_methods_KL_joint}, we show how the objective separates into two KL divergence terms:
\begin{align}
    D_\text{KL}\left( q(\bx|\bz) q(\bz) \| p\indtarg(\bx) r(\bz|\bx) \right) &= - \mathbb{E}_{q(\bz)}\left[  \mathbb{E}_{q(\bx|\bz)}\left[ \log \frac{ p\indtarg(\bx) r\indphi(\bz |\bx)}{q(\bz|\bx) q(\bx)}  \right] \right] \nonumber \\
    &= -\mathbb{E}_{q(\bx)}\left[ \log \frac{p\indtarg(\bx)}{q(\bx)}\right] - \mathbb{E}_{q(\bz)q(\bx|\bz)}\left[ \log \frac{r\indphi(\bz|\bx)}{q(\bz|\bx)} \right]  \nonumber \\
    &=   D_\text{KL}\left( q(\bx)\| p\indtarg(\bx) \right) + D_\text{KL}\left( q(\bz|\bx) \| r\indphi(\bz |\bx) \right) \nonumber \\
    &\geq D_\text{KL}\left( q(\bx)\| p\indtarg(\bx) \right)
    \label{eqn:ch5_methods_KL_based_derivation}
\end{align}
As mentioned earlier, the lower bound on $\mathcal{L}(\bphi,\btheta)$ or upper bound on \linebreak $D_\text{KL}\left( q\indthetacf(\bx|\bz) q\indthetacc(\bz) \| p\indtarg(\bx) r(\bz|\bx) \right)$ becomes tight when $r(\bz|\bx;\bphi^{\text{opt}}) = q(\bz|\bx)$, which is \eqqref{eqn:ch5_methods_KL_based_derivation}. Suboptimal $\bphi$ imply  bounds on the objective owing to the positivity of $D_\text{KL}\left( q(\bz|\bx) \| r\indphi(\bz |\bx) \right)\geq 0$.

The advantage of the proposed method for identifying CVs and learning a predictive coarse-graining model becomes clearer when we directly utilize the reference potential energy $U(\bx)$ (which we assume to be available in this paper). The objective $\mathcal{L}(\bphi, \btheta)$, which is the negative KL divergence defined by the joint distributions, is subject to maximization with respect to the parameters $\btheta$ and $\bphi$:
\begin{align}
 \mathcal{L}(\bphi, \btheta) & =  -D_\text{KL} \left( q\indthetacc(\bz) q\indthetacf(\bx| \bz) \| p\indtarg(\bx) r\indphi(\bz|\bx) \right) \nonumber  \\
 & = \left\langle \log p\indtarg(\bx) \right\rangle_{q\indtheta(\bx,\bz)} +\left\langle \log \frac{r\indphi(\bz|\bx) }{q\indthetacc(\bz) q\indthetacf(\bx| \bz)} \right\rangle_{q\indtheta(\bz, \bx)} \nonumber \\
 & = -\beta \left\langle U(\bx)\right\rangle_{q\indtheta(\bx,\bz)} +\left\langle \log \frac{r_{\bphi}(\bz|\bx) }{q\indthetacc(\bz) q\indthetacf(\bx| \bz)} \right\rangle_{q\indtheta(\bz, \bx)}
 \label{eqn:ch5_methods_kl_physics}
\end{align}
Maximizing \eqqref{eqn:ch5_methods_kl_physics} solely involves expectations with respect to the generative model, from which it is easy to draw samples from. Explicitly there are no expectations with respect to the target density $p\indtarg(\bx)$, which would require an approximation with data. Instead of data, the target density $p\indtarg(\bx)$ contributes to the learning of the parameters  $(\bphi, \btheta)$ through the interatomic potential energy $U(\bx)$ assessed for samples of the generative model $q(\bx|\bz)q(\bz)$. Note that the normalization constant of $p\indtarg(\bx) $ is independent of $\bphi$ and $\btheta$ and has been omitted in \eqqref{eqn:ch5_methods_kl_physics}.
We are aware that the method requires a potential energy function $U(\bx)$, which can be assessed at $\bx$. This is always the case for systems where we can set up MD or MCMC simulations, although we do circumvent the need to simulate a trajectory or draw reference samples by directly incorporating the available physics expressed by the potential energy.

\subsection{Reverse or forward KL divergence?}
\label{sec:ch5_methods_comparison_var_data}

In the following, we point out commonalities and differences between the proposed approach relying on the reverse KL divergence as introduced in \eqqref{eqn:ch5_method_reverse_kl} and the forward KL divergence  (\eqqref{eqn:ch5_method_forward_kl}). The latter has been successfully employed for the development of coarse-graining methodologies \cite{hernandez2017_vae, noe2018_autoencoder_timelagged, mohamed2018_vae} and with a focus on CV discovery in combination with predictive coarse-graining in \cite{schoeberl2019_pcvs}.

The data-driven objective is based on minimizing the following KL divergence:
\begin{equation}
D_\text{KL}\left( p\indtarg(\bx) \| q\indtheta(\bx) \right).
\label{eqn:ch5_methods_compare_forward_KL}
\end{equation}
Reformulating the minimization of \eqqref{eqn:ch5_methods_compare_forward_KL} to a maximization problem, the lower bound based on the summation over terms corresponding to each datum $\bxi$ of a set of $ \bx^\data = \left\{ \bxi \right\}_{i=1}^N$ is written as:
\begin{align}
	\mathcal L^{\text{forward}}(\btheta, \bphi; \bx^\data) = &\sum_{i=1}^N \mathbb E_{r\indphi(\bz^\ibrack|\bxi)} \left[ - \log r\indphi(\bz^\ibrack|\bxi)  + \log q\indtheta(\bxi,\bz^\ibrack) \right] \nonumber \\
	= &-\sum_{i=1}^N D_\text{KL}\left(r\indphi(\bz^\ibrack|\bxi) \| q\indtheta(\bz^\ibrack)\right) \nonumber \\
	&+ \sum_{i=1}^N \mathbb E_{r\indphi(\bz^\ibrack|\bxi)} \left[\log q\indtheta(\bxi|\bz^\ibrack) \right].
	\label{eqn:ch5_methods_compare_data_lower_bound}
\end{align}
The objective above depicts the lower bound on the marginal log-likelihood and has been constructed in the context of data-driven variational inference \cite{beal2003_variational_inference, kingma2014_autoencoding_vb, ranganath2014_bbvi}. The first component in \eqqref{eqn:ch5_methods_compare_data_lower_bound} implies minimizing   $D_\text{KL}\left(r\indphi(\bz^\ibrack|\bxi) \| q\indtheta(\bz^\ibrack)\right)$ in an aggregation of all considered $\bxi$. Hence, in aggregation the pre-images of $\bxi$, expressed by the approximate posterior, should resemble the generative component $q\indtheta(\bz)$, whereas the latter term in \eqqref{eqn:ch5_methods_compare_data_lower_bound} accounts for the reconstruction loss of encoded  pre-images $\bz^\ibrack$ (encoded through $r\indphi(\bz^\ibrack|\bxi)$) to its origin $\bxi$).

Minimizing the reverse KL divergence as introduced in \eqqref{eqn:ch5_method_reverse_kl} with
\begin{equation*}
    D_\text{KL}\left( q\indtheta(\bx)  \|   p\indtarg(\bx) \right)
\end{equation*}
implies a tractable maximization with respect to $(\bphi, \btheta)$ of the following objective based on \cite{ranganath2016_hvm}:
\begin{equation}
 \mathcal{L}(\bphi, \btheta) =  \underbrace{-\beta \left\langle U(\bx)\right\rangle_{q\indtheta(\bx,\bz)}}_{\ast} + \underbrace{\mathbb{E}_{q\indtheta(\bx,\bz)} \left[ \log r\indphi(\bz|\bx) \right]}_{\dag} + \underbrace{\mathbb{H}(q\indtheta(\bx,\bz))}_{\ddag}.
 \label{eqn:ch5_methods_methods_compare_kl_physics}
\end{equation}
We comment on the meaning of the indicated terms in \eqqref{eqn:ch5_methods_methods_compare_kl_physics}; however, note that the optimization always needs to be regarded in the composition of all terms.
\begin{itemize}
\item[$\ast$)] Maximizing $\mathcal{L}(\bphi, \btheta) $ seeks to minimize $\beta \left\langle U(\bx)\right\rangle_{q\indtheta(\bx,\bz)}$, which corresponds to the average potential energy of the system evaluated under the generative model $q\indtheta(\bx)$.
\item[$\dag$)] Maximize the expected log-probability that a given fine-scale realization $\bxi$ (with the corresponding latent pre-image $\bzi$ ) drawn from the joint of the generative model, $q\indtheta(\bx,\bz)$, can be reconstructed by $r\indphi(\bz|\bx)$ based on $\bxi$.
\item[$\ddag$)] Maximize the entropy of the generative model $\mathbb{H}(q\indtheta(\bx,\bz))$.
\end{itemize}
Note that all aforementioned contributions must be seen in the composition, and maximizing $\mathcal{L}(\bphi, \btheta) $ with respect to $(\bphi, \btheta)$ maximizes the balance of all.
Most important is that the involved objective in the reverse KL divergence does not encompass any expectations with respect to $p\indtarg(\bx)$, which need to be approximated by data as is the case in $\mathcal L^{\text{forward}}(\btheta, \bphi; \bx^\data)$.

We discuss in the next section the particulars of the optimization with respect to $\btheta,\bphi$, and also specify the form of the densities involved, i.e., $q\indtheta$ and $r\indphi$.

\subsection{Model specification and gradient derivation}
\label{sec:ch5_methods_model_gradient}

In the sequel we introduce a general approach for parametrizing distributions $q\indtheta(\bx,\bz)$ and $r\indphi(\bz|\bx)$ and provide an approach for optimizing parameters with variance-reduction methods, enabling accelerated convergence.

\subsubsection{Model specification}
\label{sec:ch5_methods_model}

We base the model specification on previous work in the context of data-driven CVs discovery \cite{schoeberl2019_pcvs}. The model involves two components, ($q(\bx|\bz)$ and $q(\bz)$), with respect to the generative path and the encoder  $r(\bz|\bx)$ in the recognition path.

As we seek to obtain a set of lower-dimensional coordinates representing characteristic and slow coordinates of the system, we aim to provide complexity in the mapping and thus the encoder and decoder components $r(\bz|\bx)$ and $q(\bx|\bz)$, respectively, and simple descriptions of the CVs through $q(\bz)$. Pushing complexity to the involved mappings and assuming simple correlations in $q(\bz)$ yields CVs capturing the most relevant features of the atomistic system compressed in low dimensions \cite{amadei1993, ferguson2011b}.

The distribution $q\indthetacc(\bz)$, which the obtained CVs are supposed to follow and which we desire to be simple, is represented as a standard Gaussian with unit diagonal variance:
\be
q\indthetacc(\bz) = q(\bz) = \mathcal N (\bz; \boldsymbol{0}, \boldsymbol{I}).
\label{eqn:ch5_methods_model_prior}
\ee
The simplicity induced by \eqqref{eqn:ch5_methods_model_prior} is balanced by employing a flexible mapping given latent CVs $\bz$ to fine-scale atomistic coordinates $\bx$ (probabilistic decoder) with
\begin{equation}
	\label{eqn:ch5_methods_model_map}
	q\indthetacf(\bx|\bz) = \mathcal N\left(\bx; \bmu\indthetacf(\bz), \bm{S}\indthetacf\right),
\end{equation}
where the nonlinear mapping
\be
\bmu\indthetacf(\bz) = f\indthetacf^{\bmu}(\bz),
\label{eqn:ch5_methods_model_f_decoder_mapping}
\ee
with $\bz \mapsto f\indthetacf^{\bmu}(\bz)$ ($f\indthetacf^{\bmu} : \mathbb R^{n\indcc} \mapsto \mathbb R^{n\indf}$) is expressed by a flexible (multilayer) neural network \cite{rumelhart1986, malsburg1986, haykin1998}.
The Gaussian in  \eqqref{eqn:ch5_methods_model_map} with the flexible mean $\bmu\indthetacf(\bz)$  is then fully defined by considering a 
diagonal covariance matrix with $\bm{S}\indthetacf=\mathrm{diag}(\bsig^2_{\btheta\indcf})$ \cite{mattei2018}.
We omit the subscripts of $\btheta$, as the latent generator $q(\bz)$ does not depend on parameters. Thus, we write $\btheta = \btheta\indthetacf$.
We treat the entries $\sigma_{\btheta, j}^2$ directly as parameters without dependence on latent CVs $\bz$. Maintaining $\sigma^2_{\btheta,j} > 0$ is ensured by optimizing $\log \sigma^2_{\btheta,j}$ instead.

In a similar fashion, compared to the model of $q\indthetacf(\bx|\bz)$, we express the encoder that approximates the actual posterior distribution $p(\bz|\bx)$ as follows:
\begin{equation}
	r\indphi(\bz|\bx) = \mathcal N \left(\bz; \bmu\indphi(\bx), \bm{S}\indphi(\bx)\right),
	\label{eqn:ch5_methods_model_f_encoder_mapping}
\end{equation}
with the diagonal covariance matrix  $\bm{S}\indphi(\bx)=\mathrm{diag}\left(\bsig\indphi^2(\bx)\right)$. Likewise, 
 $\bmu\indphi(\bx)$ and $\log \bsig_{\bphi}^2(\bx)$ are obtained from encoding neural networks $f^{\bmu}\indphi(\bx)$ and $f^{\bsig}\indphi(\bx)$, respectively:
\begin{equation}
	\bmu\indphi(\bx) = f\indphi^{\bmu}(\bx) \quad \text{and} \quad \log \bsig\indphi^2(\bx)= f\indphi^{\bsig}(\bx).
	\label{eqn:fphi}
\end{equation}
The actual but intractable posterior $q(\bz|\bx)$ will differ from a multivariate normal distribution, for which we compensate by providing a flexible mean in $r\indphi(\bz|\bx)$. Structural correlations revealed by a full rank covariance matrix represent an interesting avenue to be explored \cite{pinheiro1996}; however, this is not part of this paper. The employed models resemble those developed earlier in the context of CV discovery. Therefore, we refer to the discussion in \cite{schoeberl2019_pcvs} justifying the use of the neural networks.

We utilize the following general structure for the decoding neural network $f\indthetacf^{\bmu}(\bz)$:
\be
\label{eqn:ch5_methods_decoding_net}
f\indtheta^{\bmu, K_q}(\bz) =\left( l\indthetacf^{(K_q+1)} \circ \tilde{a}^{(K_q)} \circ l\indtheta^{(K_q)} \circ \cdots \circ \tilde{a}^{(1)} \circ l\indtheta^{(1)} \right)(\bz).
\ee
with $K_q$ hidden layers. In a similar manner, we define the encoding networks for $\bmu\indphi(\bx)$ and $\bsig^2\indphi(\bx)$  of $r\indphi(\bz|\bx)$:
\be
\label{eqn:ch5_methods_encoding_net_shared}
f^{K_r}\indphi(\bx) = \left( a^{(K_r)} \circ l\indphi^{(K_r)} \circ \cdots  \circ a^{(1)} \circ  l\indphi^{(1)} \right)(\bx),
\ee
which leads to the mean and diagonal terms of the covariance matrix with
\be
\label{eqn:ch5_methods_encoding_mu_sig}
f^{\bmu}\indphi(\bx) = l\indphi^{(K_r+1)}\left(f^{K_r}\indphi(\bx) \right) \quad \text{and} \quad 
f^{\bsig}\indphi(\bx) = l\indphi^{(K_r+2)}\left(f^{K_r}\indphi(\bx) \right).
\ee
The linear layers used in the above expressions are denoted as $l^{(i)}$, e.g., mapping a  variable $\boldsymbol{y}$ to the output with $l^{(i)}(\boldsymbol{y}) = \boldsymbol{W}^{(i)} \boldsymbol{y} + \boldsymbol{b}^{(i)}$. The nonlinearities in  $f^{(\cdot)}_{(\cdot)}$ are implied by activation $a(\cdot)$. Encoding and decoding functions are indicated by the superscripts $\bphi$ and $\btheta$, respectively.  Activation functions belonging to the encoder are $a^{(i)}$, and those involved in decoding $\bz$ are  $\tilde{a}^{(i)}$.
The size of $\boldsymbol{W}^{(i)}$ is specified by the input dimension, which could be the output of a precedent layer $l^{(i-1)}(\boldsymbol{y})$, and  the output dimension, which we specify with  $d_{l^{(i)}}$. This leads to a matrix $\boldsymbol{W}^{(i)} $ of  dimension $ d_{l^{(i)}} \times d_{l^{(i-1)}}$.
The corresponding parametrization details with depth $K$ and activations of the networks are specified with the corresponding numerical illustrations in \secref{sec:ch5_nummerical_illustrations}.

\subsubsection{Gradient computation and reparametrization}
\label{sec:ch5_methods_gradient}

This section is devoted to deriving relevant gradients of the objective  $\mathcal{L}(\bphi, \btheta)$ in \eqqref{eqn:ch5_methods_elbo_w_bound}, which involve the fine-scale potential energy $U(\bx)$. We show a noise-reducing gradient estimator by utilizing reparametrization \cite{kingma2014_welling_gradient, rezende2014_stoch_backpropagation}.

The focus is on the first component in  \eqqref{eqn:ch5_methods_elbo_w_bound}, which depends only on the parameters $\btheta$. We write for the corresponding derivative:
\begin{align}
    -\beta \frac{\partial}{\partial \btheta} \left\langle U(\bx) \right\rangle_{q\indtheta(\bx|\bz)} =& -\beta \frac{\partial}{\partial \btheta}  \int \int q\indtheta(\bx|\bz) q(\bz) U(\bx) ~d\bx ~d\bz \nonumber \\
    = &-\beta  \int q(\bz) \underbrace{\hleq{ \frac{\partial}{\partial \btheta} \left( \int q\indtheta(\bx |\bz)  U(\bx) ~d\bx \right)}}_{\nabla\indtheta\mathbb{E}_{q\indtheta(\bx |\bz)}\left[  U(\bx) \right]} ~d\bz.
    \label{eqn:ch5_methods_grad_u_x}
\end{align}
In the last line of the above equation, we note the expression $\nabla\indtheta\mathbb{E}_{q\indtheta(\bx |\bz)}\left[  U(\bx) \right]$; this is for the case of using approximate MC estimators, highly affected by noise, as discussed in \cite{ranganath2014_bbvi}. This would hamper the optimization even when employing stochastic techniques. The variance of the approximate estimator of $\nabla\indtheta\mathbb{E}_{q\indtheta(\bx |\bz)}\left[  U(\bx) \right]$ can be reduced by the reparametrization of $q\indtheta(\bx |\bz)$. This is done by introducing an auxiliary random variable $\beps$, which gives rise to $\bx$ by a differentiable transformation:
\begin{equation}
\bx = g\indtheta(\beps; \bz) ~ \text{with} ~ \beps \sim p(\beps).
\label{eqn:ch5_methods_repar_method}
\end{equation}
With the mapping, $g\indtheta : \beps \rightarrow \bz$, the following holds by change of variables:
\be
q\indtheta(\bx|\bz) = p\big(g\indtheta^{-1}(\bx; \bz)\big) \bigg\lvert \frac{\partial g\indtheta^{-1}(\bx; \bz)}{\partial \bx}\bigg\rvert,
\label{eqn:ch5_methods_change_of_variables}
\ee
where the inverse function of $g\indtheta$,  $g\indtheta^{-1}: \bx \rightarrow \beps$ leads to  $\beps = g\indphi^{-1}(\bx; \bz)$. Different possibilities of auxiliary distributions and invertible transformations are discussed in more detail in \cite{ruiz2016generalized}. With the introduced transformation, we can rewrite the derivative with:
\begin{align}
\nabla\indtheta\mathbb{E}_{q\indtheta(\bx |\bz)} \left[  U(\bx) \right]
&= \mathbb E_{p(\beps)} \left[\nabla_{\btheta} U\left(g\indtheta(\beps; \bz)\right) \right] \nonumber \\ 
&= \mathbb E_{p(\beps)} \left[ \frac{ \partial U\left(g\indtheta(\beps; \bz)\right)}{\partial \bx} \frac{\partial g\indtheta(\beps; \bz))}{\partial \btheta} \right].
\label{eqn:ch5_methods_reparametrized}
\end{align}
The auxiliary random variables $\beps$ follow a Gaussian with $\beps^{(l)} \sim p(\beps) = \mathcal N( \boldsymbol{0}, \boldsymbol{I})$. The corresponding transformation for representing the random variables $\bx $ is:
\begin{equation}
 \bx = g\indtheta(\beps; \bz) = \bmu\indtheta(\bz) + \bsig\indtheta(\bz) \odot \beps.
\end{equation}
Replacing the expression $\nabla\indtheta\mathbb{E}_{q\indtheta(\bx |\bz)}\left[  U(\bx) \right]$ in \eqqref{eqn:ch5_methods_grad_u_x} with \eqqref{eqn:ch5_methods_reparametrized} leads to:
\begin{align}
    -\beta \frac{\partial}{\partial \btheta} \left\langle U(\bx) \right\rangle_{q\indtheta(\bx|\bz)} &= -\beta \left\langle  \mathbb E_{p(\beps)} \left[ \frac{ \partial U\left(g\indtheta(\beps; \bz)\right)}{\partial \bx} \frac{\partial g\indtheta(\beps; \bz))}{\partial \btheta} \right] \right\rangle_{q(\bz)} \nonumber \\
    &= -\beta \left\langle \underbrace{\hleq{ \frac{ \partial  U\left(g\indtheta(\beps; \bz)\right)}{\partial \bx} }}_{=-\mathbf{F}(\bx)} \frac{\partial g\indtheta(\beps; \bz))}{\partial \btheta}\right\rangle_{p(\beps) q(\bz)}.
    \label{eqn:ch5_method_force_gradient}
\end{align}
First, the physically less interesting part in \eqqref{eqn:ch5_method_force_gradient}  is the  contribution $ \frac{\partial g\indtheta(\beps; \bz))}{\partial \btheta}$, which can be estimated by employing efficient backpropagation and automatic differentiation algorithms for neural networks \cite{rumelhart1986_back_prop, rumelhart1986}.
However, the more physically relevant component, the gradient of the atomistic potential,  $\frac{ \partial  U\left(g\indtheta(\beps; \bz)\right)}{\partial \bx}$, is involved in \eqqref{eqn:ch5_method_force_gradient}. The gradient of the potential $U(\bx)$ with respect to $\bx$ equals the negative interatomic force $\mathbf{F}(\bx)$, evaluated at $\bx$, where $\bx = g\indtheta(\beps; \bz)$. This latter term incorporates physics into the gradient computation of $\mathcal{L}(\bphi, \btheta)$ in the form of interatomic forces. This is  the source from which physics are embedded into our proposed model and drives the optimization of $\mathcal{L}(\bphi, \btheta)$ by querying the force field at samples of $q\indtheta(\bx)$. Notably, the forces are incorporated at atomistic positions $\bx$, which are determined by sampling as follows.
\begin{enumerate}[label=(\roman*)]
    \item Draw a sample from the generative distributions: $\bzi \sim q(\bz)$ which is simple to sample from.
    \item Then obtain a sample from the auxiliary distribution: $\beps^{(j)} \sim p(\beps)$.
    \item Determine the corresponding atomistic representation of $(\bzi, \beps^{(j)})$ with:
    $\bx^{(i,j)} =  g\indtheta(\beps^{(j)}; \bzi) =  \bmu\indtheta(\bzi) + \bsig\indtheta(\bzi) \odot \beps^{(j)}$.
\end{enumerate}
This means we evaluate the force $\mathbf{F}$ at samples $\bx^{(i,j)}$; no reference data are required in this process. 

The force evaluation at atomistic coordinates $\bx$ is the heart of common MD software such as LAMMPS \cite{plimpton1995_lammps}, GROMACS \cite{berendsen1995, lindahl2001, spoel2005, hess2008, sander2013, prall2015, abraham2015}, and OpenMM \cite{eastman2017_openmm}. The MD simulators are highly sophisticated in terms of efficiency and allow us to employ this optimized force evaluation function in our development.

In this work, we develop a PyTorch module that incorporates OpenMM \cite{eastman2017_openmm} in the backward pass, which enables efficient optimization by querying the forces computed by  OpenMM at input positions governed by $q\indtheta(\bx)$. We are continuously developing the software on GPU platforms, and it will be made available\footnote{Software available upon publication on \url{https://github.com/mjs.../...}.}.

\subsection{Training}
\label{sec:ch5_methods_training}

Training the model parameters $(\bphi, \btheta)$ requires some attention as variational models tend to be mode-focusing \cite{bishop2006, shu2018_amortized_inference}. If parameters update too rapidly, in terms of configurations of $p\indtarg(\bx)$ that have been explored by $q\indtheta(\bx)$ thus far, relevant conformations could be missed. However, compared with the data-driven approach, the proposed variational coarse-graining methodology offers strategies ensuring that relevant conformations are captured and incorporates querying of the potential $U(\bx)$ into the learning procedure. In data-driven schemes, once the data is obtained, there is no control on exploring unobserved conformations \cite{sittel2018_cv_nonlinear_map, ferguson2011b}.
Remedy, next to employing stochastic optimization with adaptive step size control \cite{spall2003_stochastic_optimization}, provide tempering approaches \cite{marinari1992_tempering, chandra2018_tempering_nn}. These start at high initial temperatures or low inverse temperatures, with, e.g., $0 \leq \beta_1$, whereas $\beta = 0$ resembles a uniform target distribution. A sequence of $K$ temperatures and related inverse temperatures  $0\leq \beta_1  \dots \leq  \beta_k \leq \dots \beta_K$ yields a sequence of target distributions with  \cite{doucet2001_smc, delmoral2006_smc, delmoral2011_smc}
\begin{equation}
    p\indtarg(\bx;\beta_k) = \frac{1}{\mathrm{Z}(\beta_k)} e^{-\beta_k U(\bx)} , \quad \forall ~ k  ~\in \{1, \dots, K\},
\end{equation}
while $\beta_K$ equals the target simulation temperature $\beta\indtarg$.

Instead of directly minimizing $D_\text{KL}\left(q(\bx) \| p\indtarg(\bx;\beta\indtarg)\right)$,
we minimize subsequent $D_\text{KL}\left(q(\bx) \| p\indtarg(\bx;\beta_\mathbf{k})\right)$ while we obtain optimal $(\bphi_k,\btheta_k)$,  which we use as initial parameters for minimizing  $D_\text{KL}\left(q(\bx) \| p\indtarg(\bx;\beta_{\mathbf{k+1}})\right)$. However, the size of the increment between two subsequent temperature steps $\Delta \beta_k = \beta_{k+1} - \beta_k $ is a difficult choice.

Therefore, we develop an adaptive scheme for gradually increasing $\beta_k$, which adjusts the proposed $\Delta \beta_k$ such that the relative difference in subsequent KL divergences estimated at $\beta_k$ and $\beta_{k+1}$ does not exceed a threshold $c_\text{max}$.
We define the relative increase of the KL divergence between $\beta_k$ and $\beta_{k+1}$ with:
\be
\frac{D_\text{KL}\left(q(\bx) \| p\indtarg(\bx;\beta_{k+1})\right) - D_\text{KL}\left(q(\bx) \| p\indtarg(\bx;\beta_{k})\right) }{D_\text{KL}\left(q(\bx) \| p\indtarg(\bx;\beta_{k})\right)}.
\label{eqn:ch5_methods_rel_kl_inc}
\ee
By employing the derived upper bound on the KL divergence, which is defined in \eqqref{eqn:ch5_methods_kl_physics}, we can rewrite \eqqref{eqn:ch5_methods_rel_kl_inc}
as
\be
c_k = \frac{\log(\mathrm{Z}(\beta_{k+1})) - \log(\mathrm{Z}(\beta_{k})) + (\beta_{k+1} - \beta_{k}) \left\langle U(\bx) \right\rangle_{q(\bx,\bz)}}{\log \mathrm{Z}(\beta_k) + \beta_k\left\langle U(\bx) \right\rangle_{q(\bx,\bz)} - \left\langle \log r(\bz|\bx) \right\rangle_{q(\bx,\bz)} - \mathbb{H}(q(\bx,\bz))}.
\label{eqn:ch5_methods_rel_kl_inc_ext}
\ee
Besides the (log-)difference of the normalization constants, $\log(\mathrm{Z}(\beta_{i+1})) - \log(\mathrm{Z}(\beta_{i}))$, and $\log(\mathrm{Z}(\beta_{k})$, all remaining components in \eqqref{eqn:ch5_methods_rel_kl_inc_ext} are accessible through MC estimators. The supporting material in Appendix \ref{sec:ch5_appendix_si_est_kl_increase} includes an approximation of $\log(\mathrm{Z}(\beta_{k+1})) - \log(\mathrm{Z}(\beta_{k})$ and $\log(\mathrm{Z}(\beta_{k})$.
The procedure for updating the temperature is summarized in Algorithm \ref{alg:ch5_method_kl_tempering}.

\begin{algorithm}[H]
\label{alg:ch5_method_kl_tempering}
		\KwIn{Converged model with its parameter $(\bphi_k, \btheta_k)$ at current inverse temperature $\beta_k$; $c_\text{max}$, maximal relative increase in $D_\text{KL}$; $\Delta \beta_\text{max}$, the temperature increment; Current step $k$.}
		\KwOut{$\beta_{k+1}$}
		\textbf{Initialize:} $s:=0$, $f_s := 1.0$. \;
		\While{$c^s_k > c_\text{max}$}{
		\nonl \textbf{Propose new inverse temperature $\beta^s_{k+1}$:} \;
		$\beta^s_{k+1} = \beta_k + f_s \Delta \beta_\text{max}$ \;
		\nonl \textbf{Estimate rel. increase $c^s_k$ with proposed $\beta^s_{k+1}$:} \;
		See computation in \eqqref{eqn:ch5_methods_rel_kl_inc_ext}. \;
		\nonl \textbf{Update $f_{s}$ for proposing a new maximal increase in $\beta$:} \;
		$f_{s+1} = 0.6 f_{s}$ \;
		\nonl \textbf{Update step:} \;
		$s = s + 1$.\;
		}
		\textbf{Set:} $\beta_{k+1} = \beta^s_{k+1}$ \;
		\textbf{Update:} $k = k+1$ \;
		\textbf{Continue optimization with:} $\log  p\indtarg (\bx;\beta_{k}) \propto e^{\beta_{k} U(\bx)}$ \;
	\caption{Tempering scheme for updating $\beta_k$. We set $\Delta \beta_\text{max} = \num{1.e-3}$ and $c_\text{max} = 1.0$.}
\end{algorithm}

\section{Numerical illustrations}
\label{sec:ch5_nummerical_illustrations}

The following section demonstrates the developed methodology based on a double well potential in \secref{sec:ch5_nummerical_illustrations_double_well} and an alanine dipeptide in \secref{sec:ch5_nummerical_illustrations_ala2}.

\subsection{Double well}
\label{sec:ch5_nummerical_illustrations_double_well}

\begin{figure}
    \centering
    \includegraphics[width=0.7\textwidth]{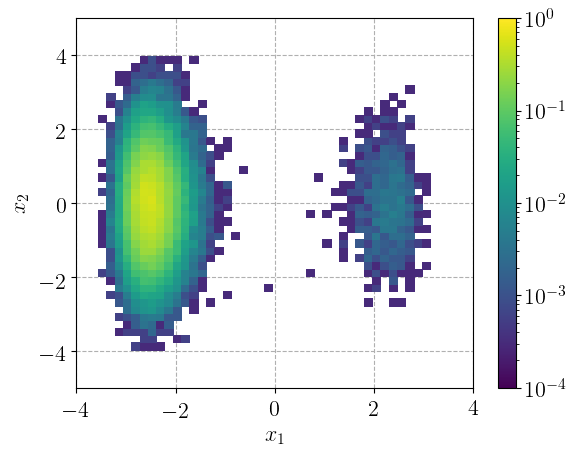}
    \caption{Reference potential energy $U(\bx)$. The color relates to the value of $U(\bx)$ quantified by the \emph{logarithmic} color bar on the right. Most MCMC random walk approaches will discover only one of the depicted potential energy basins.}
    \label{fig:ch5_numill_doublewell_reference_pot}
\end{figure}

This section shows the capabilities of the proposed method in the context of a two-dimensional double well potential energy function $U(\bx)$ ($\dim(\bx)=2$) that exhibits two distinct modes distinguishable in the $x_1$ direction. One of the modes is favorably explored owing to its lower potential energy. The potential is quadratic in the $x_2$ direction, as depicted in \figref{fig:ch5_numill_doublewell_reference_pot}:
\begin{equation}
    U(\bx) = \frac{1}{4} x_1^4 - 3 \cdot x_1^2 + x_1 + \frac{1}{2} x_2^2.
    \label{eqn:ch5_numill_double_well_pot}
\end{equation}
The double well potential in \eqqref{eqn:ch5_numill_double_well_pot} and the  implied target distribution \linebreak $p\indtarg(\bx;\beta=1) \propto e^{-\beta U(\bx)}$ result in a distribution that is challenging to explore with purely random walk MCMC and without performing extensive fine-tuning of the proposal step. A test MCMC estimator, which was as fair as possible, did not discover the second mode for $x_1 >0 $ after \num{1e5} steps.
The natural CV of the potential $U(\bx)$ and thus of $p\indtarg(\bx) \propto e^{-U(\bx)}$ is the $x_1$ coordinate. The $x_1$ direction distinguishes the two modes that $p\indtarg(\bx)$ exhibits. We expect our algorithm to reveal CVs $z$ ``equal'' to $x_1$ or having high correlation with $x_1$. We put ``equal'' in quotes as we work in a probabilistic framework. The dimensionality of $\dim(\bz)$ is $1$.

The functional form and parameters have been taken from \cite{noe2019_boltzmann_gen} to ensure comparability. However, note that we seek to identify simultaneously the lower-di\-men\-sio\-nal characteristics revealing the relevant physics, encoded in CVs, and obtain a generative CG model for predictive purposes. In \cite{noe2019_boltzmann_gen}, the focus was on the generative component. The CVs utilized for learning are \emph{selected} rather than revealed from the physics. The latent CVs $\bz$ have the same dimensionality as $\bx$ owing to the use of \emph{invertible} neural networks that require $\dim(\bz) = \dim(\bx)$ \cite{dinh2017_realnvp}.

We employ the model as introduced in \secref{sec:ch5_methods_model} and define  the unspecified options such as the number of layers, layer dimensions, and activation functions used in the encoder and decoder as in Tables \ref{tab:ch5_double_well_encoder_actfct} and \ref{tab:ch5_double_well_decoder_actfct}, respectively.
\begin{table}[]
	\centering
	\begin{tabular}{|p{2cm} p{2.5cm} p{2cm} p{2cm} p{2.5cm} |}
		\hline
		Linear layer & Input dimension & Output  \linebreak dimension & Activation layer & Activation function \\
		\hline
		\hline
		$l\indphi^{(1)}$ & $\dim(\bx)=2$ & $d_1$ &  $a^{(1)}$ & SeLu\footnote{SeLu: $a(x) = 
			\begin{cases}
			\alpha (e^x -1) & \text{if } x<0\\
			x            & \text{otherwise}.
			\end{cases} $ See \cite{klambauer2017} for further details.}\\
		$l\indphi^{(2)}$ & $d_1$ & $d_2$ &  $a^{(2)}$ & SeLu \\
		$l\indphi^{(3)}$ & $d_2$ & $d_3$ &  $a^{(3)}$ & TanH  \\
		$l\indphi^{(4)}$ & $d_3$ & $\dim(\bz)$ &  None & - \\
		$l\indphi^{(5)}$ & $d_3$ & $\dim(\bz)$ &  None & - \\
		\hline
	\end{tabular}
	\caption{Network specification of the encoding neural network with $d_{\{1,2,3\}}=100$.}
	\label{tab:ch5_double_well_encoder_actfct}
\end{table}
\begin{table}[]
	\centering
	\begin{tabular}{|p{2cm} p{2.5cm} p{2cm} p{2cm} p{2.5cm} |}
		\hline
		Linear layer & Input dimension & Output \linebreak dimension & Activation layer & Activation function \\
		\hline
		\hline
		$l\indtheta^{(1)}$ & $\dim(\bz)=1$ & $d_3$ &  $\tilde{a}^{(1)}$ & Tanh \\
		$l\indtheta^{(2)}$ & $d_3$ & $d_2$ &  $\tilde{a}^{(2)}$ & Tanh \\
		$l\indtheta^{(3)}$ & $d_1$ & $\dim(\bx)$ &  None & - \\
		\hline
	\end{tabular}
	\caption{Network specification of the decoding neural network with $d_{\{1,2,3\}}$ as defined in Table \ref{tab:ch5_double_well_encoder_actfct}.}
	\label{tab:ch5_double_well_decoder_actfct}
\end{table}
To train the parameters $(\bphi, \btheta)$, we employ a tempering scheme as introduced in \secref{sec:ch5_methods_training} and specified in Algorithm \ref{alg:ch5_method_kl_tempering} with initial $\beta_0 = \num{1e-10}$, while the target is defined with $\beta_K = 1$. For all numerical illustrations, we employ ADAM stochastic optimization \cite{kingma2015_adam} with   $\alpha = 0.001$, $\beta_1= 0.9$, $\beta_2 =  0.999$, and $\epsilon_{\text{ADAM}}=\num{1.0e-8}$. The expectations with respect to $q(\bx,\bz) $ are computed based on $J= \num{1000}$ samples.

We will assess the trained model with respect to its predictive power and the obtained CVs in the following.

\subsubsection{Predictive CG model}
\label{sec:ch5_double_well_cg}

\begin{figure}[h]
	\centering
	\subfigure[Histogram of $q\indtheta(\bx)$ at $\beta \approx 0$ and of $p\indtarg(\bx)$.]{%
		\includegraphics[width=0.45\textwidth]{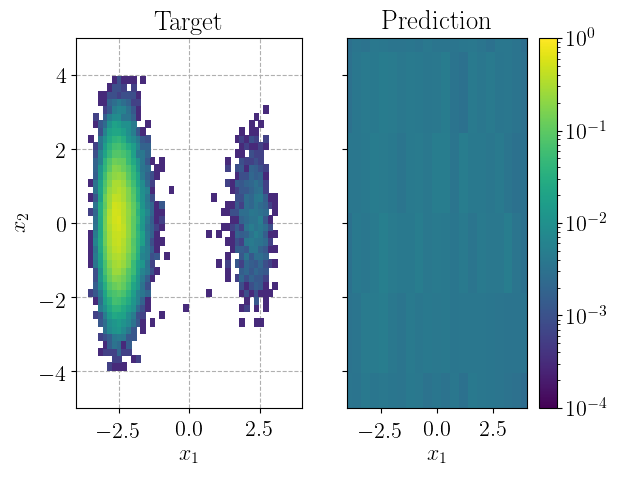}}%
	\qquad
	\subfigure[$U(x_1,x_2=0)$ at $\beta \approx 0$.]{%
		\includegraphics[width=0.45\textwidth]{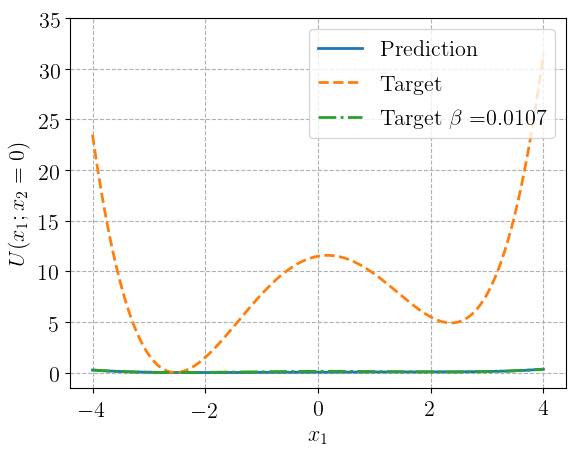}}%
	\qquad
	\subfigure[Histogram of $q\indtheta(\bx)$ at $\beta \approx 0.2$ and of $p\indtarg(\bx)$.]{%
		\includegraphics[width=0.45\textwidth]{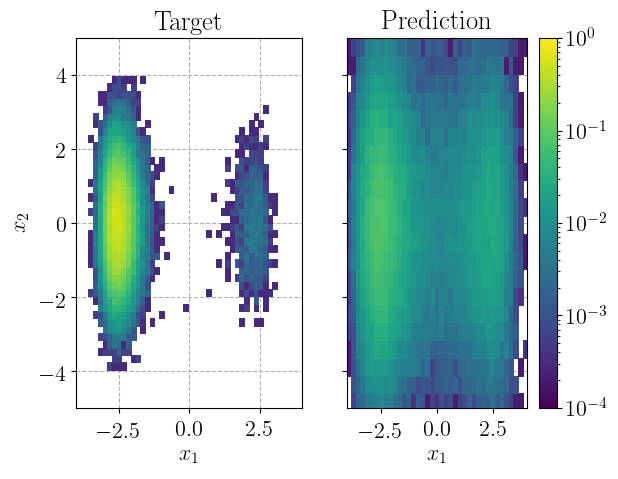}}%
	\qquad
	\subfigure[$U(x_1,x_2=0)$ at $\beta \approx 0.2$.]{%
		\includegraphics[width=0.45\textwidth]{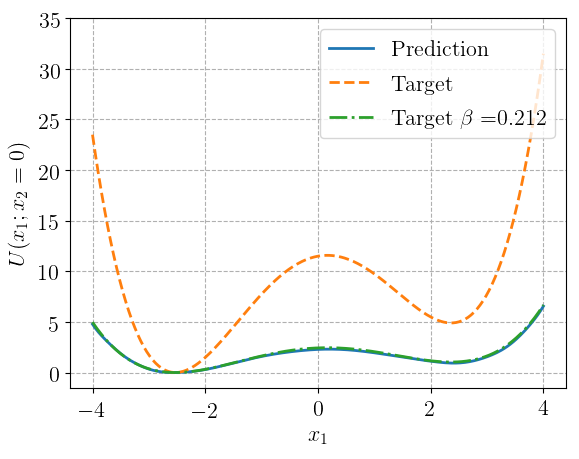}}%
	\qquad
	\subfigure[Histogram of $q\indtheta(\bx)$ at $\beta \approx 0.36$ and of $p\indtarg(\bx)$.]{%
		\includegraphics[width=0.45\textwidth]{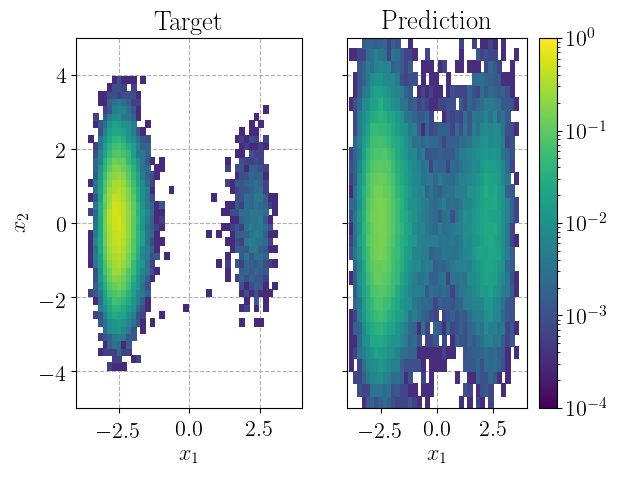}}%
	\qquad
	\subfigure[$U(x_1,x_2=0)$ at $\beta \approx 0.36$.]{%
		\includegraphics[width=0.45\textwidth]{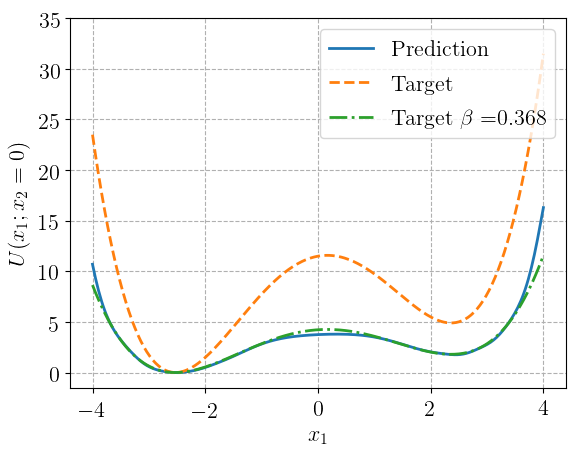}}%
	\qquad
	\caption{The left column shows histograms of the target $p\indtarg(\bx)$ (at $\beta=1$) and predictions based on $q\indphi(\bx)$ at the indicated temperature $\beta$ in the subcaptions. The right column shows a 1D slice through the potential energy $U(\bx)$ at $x_2=0$, emphasizing the two distinct modes. The figures include the reference potential for the indicated temperature $\beta_k$ with $\beta_k U(\bx)$ and an estimation of $U^\text{pred}_k(\bx)$ based on $q\indtheta(\bx)$.}
	\label{fig:ch5_double_well_prediction_potential_0}
\end{figure}

\begin{figure}
	\centering
	\subfigure[Histogram of $q\indtheta(\bx)$ at $\beta \approx 0.7$ and of $p\indtarg(\bx)$.]{%
		\includegraphics[width=0.45\textwidth]{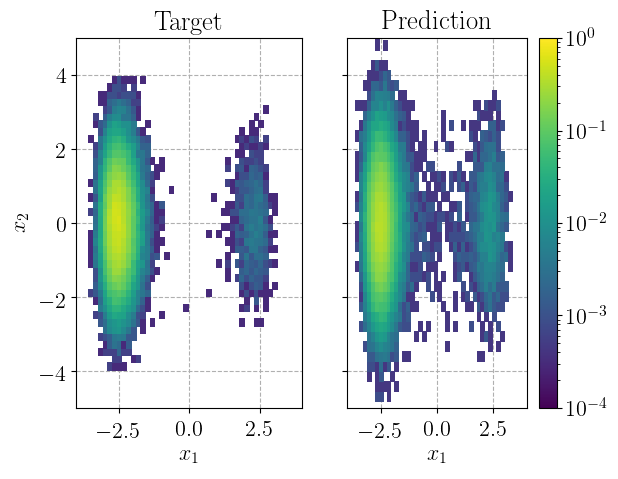}}%
	\qquad
	\subfigure[$U(x_1,x_2=0)$ at $\beta \approx 0.7$.]{%
		\includegraphics[width=0.45\textwidth]{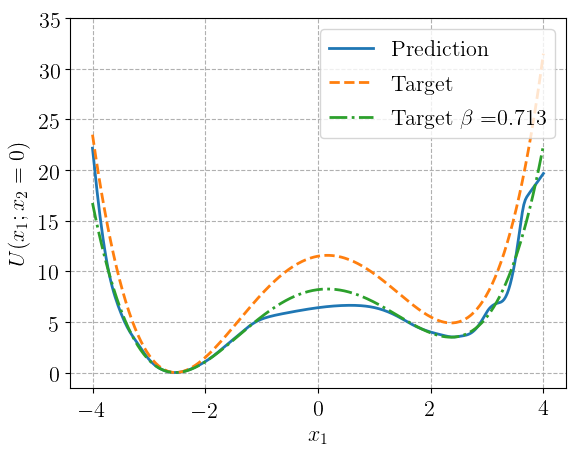}}%
	\qquad
	\subfigure[Histogram of $q\indtheta(\bx)$ at $\beta \approx 1$ and of $p\indtarg(\bx)$.]{%
		\includegraphics[width=0.45\textwidth]{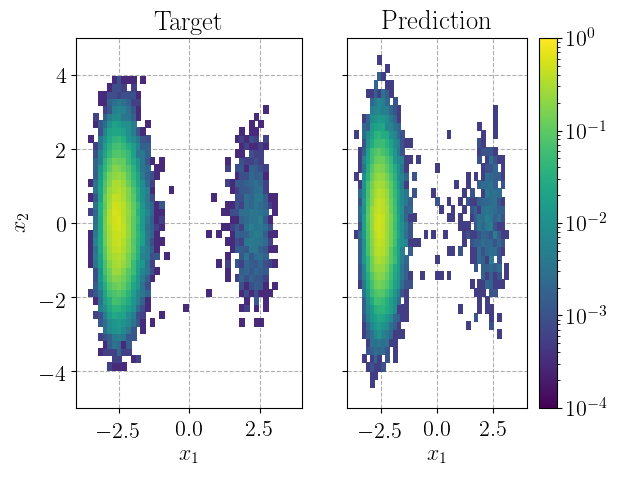}}%
	\qquad
	\subfigure[$U(x_1,x_2=0)$ at $\beta \approx 1$.]{%
		\includegraphics[width=0.45\textwidth]{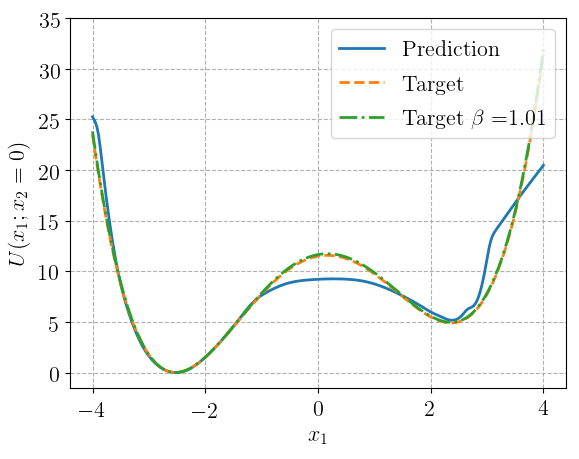}}%
	\qquad
	\caption{The left column shows histograms of the target $p\indtarg(\bx)$ (at $\beta=1$) and predictions based on $q\indphi(\bx)$ at the indicated temperature $\beta$ in the subcaptions. The right column shows a 1D slice through the potential energy $U(\bx)$ at $x_2=0$, emphasizing the two distinct modes. The figures include the reference potential for the indicated temperature $\beta_k$ with $\beta_k U(\bx)$ and an estimation of $U^\text{pred}_k(\bx)$ based on $q\indtheta(\bx)$.}
	\label{fig:ch5_double_well_prediction_potential}
\end{figure}

Figures~\ref{fig:ch5_double_well_prediction_potential_0} and \ref{fig:ch5_double_well_prediction_potential} show intermediate results obtained while training the model. The left columns depict a two-dimensional (2D) histogram containing the target histogram based on a long reference simulation obtained by employing the  Metropolis-adjusted Langevin algorithm \cite{gareth1996_mala} at $\beta=1$. Next to the histogram of $p\indtarg(\bx;\beta=1)$, we provide 2D histograms of intermediate predictions at $\beta_k$, as indicated in the sub-caption. The predictive histograms are obtained by drawing $J$ samples from the predictive distribution $q\indtheta(\bx)$. The latter is very simple and computationally efficient owing to the use of ancestral sampling \cite{liu2008} of the generative model, as explained in the \secref{sec:ch5_methods_pgm}. The right columns of Figures \ref{fig:ch5_double_well_prediction_potential_0} and \ref{fig:ch5_double_well_prediction_potential}  provide the reference potential energy $U(x_1,x_2=0)$,  the intermediate target potential $\beta_k U(x_1,x_2=0)$, and the predicted potential $U_k^{\text{pred}}(x_1,x_2=0)$ after convergence of $(\bphi, \btheta)$ at temperature $\beta_k$. For the intermediate steps, we estimate $U_k^{\text{pred}}(x_1,x_2=0)$ as follows:
\begin{equation}
    U_k^{\text{pred}}(x_1,x_2=0) \propto -\frac{1}{\beta_k} \log q\indtheta(x_1, x_2=0).
\end{equation}
We note that the evaluation of $\log q\indtheta(x_1, x_2=0)$ requires approximation of the integral $\log q\indtheta(x_1, x_2=0)= \int q(\bx|\bz) q(\bz) ~d\bz$, which induces noise. The aforementioned integral has been approximated by $N=\num{5000}$ samples drawn from $q(\bz)$.

\begin{figure}[]
    \centering
    \subfigure[Upper and lower bounds of the training objective.]{
    \includegraphics[width=0.45\textwidth]{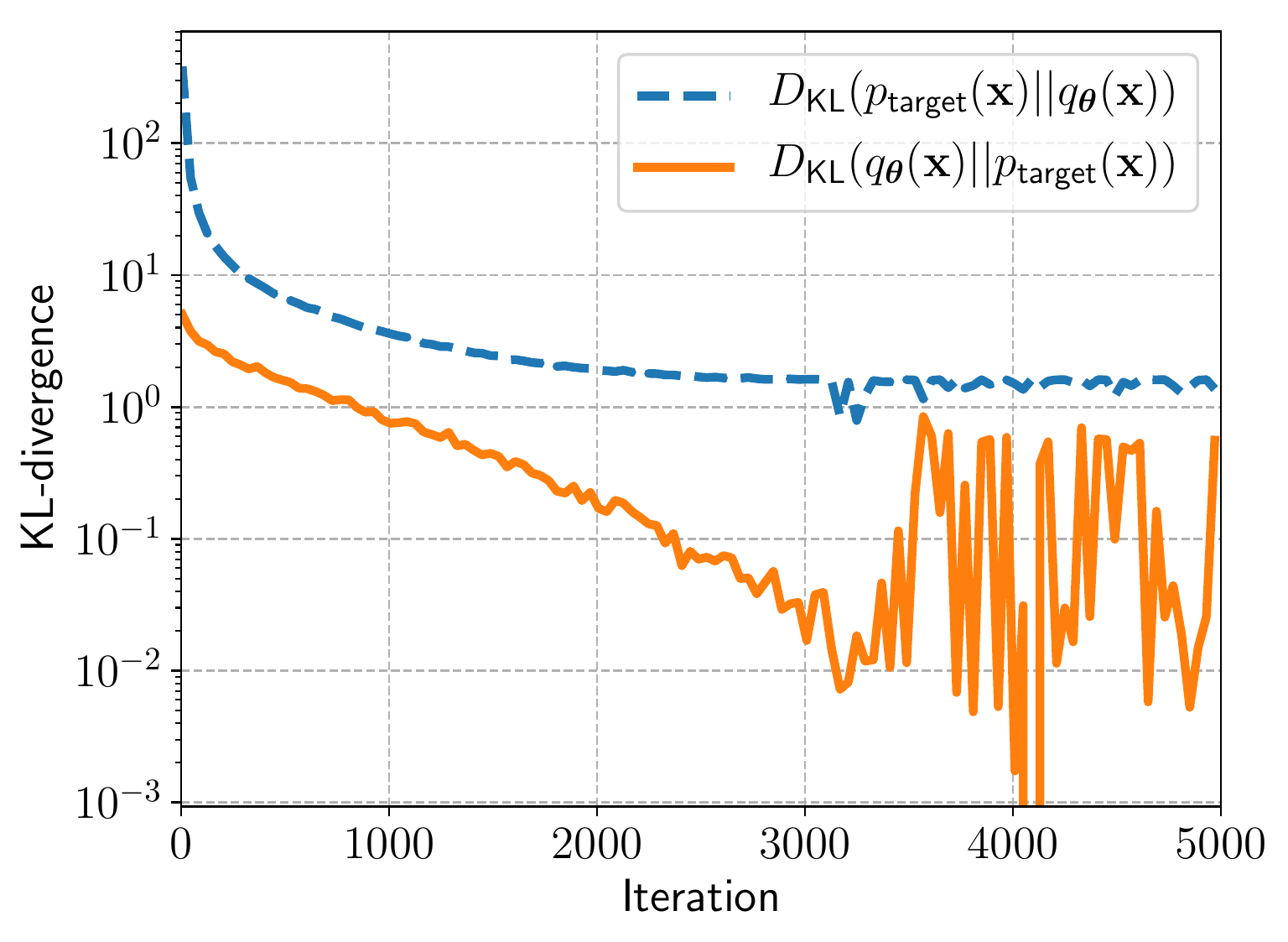}
    \label{fig:ch5_double_well_kl_bounds}}
    \qquad
    \subfigure[Predicted mean and standard deviations compared with reference data-based estimates. The subscripts of $\mu$ and $\sigma$ indicate the corresponding $x_1$ and $x_2$ directions.]{
    \includegraphics[width=0.45\textwidth]{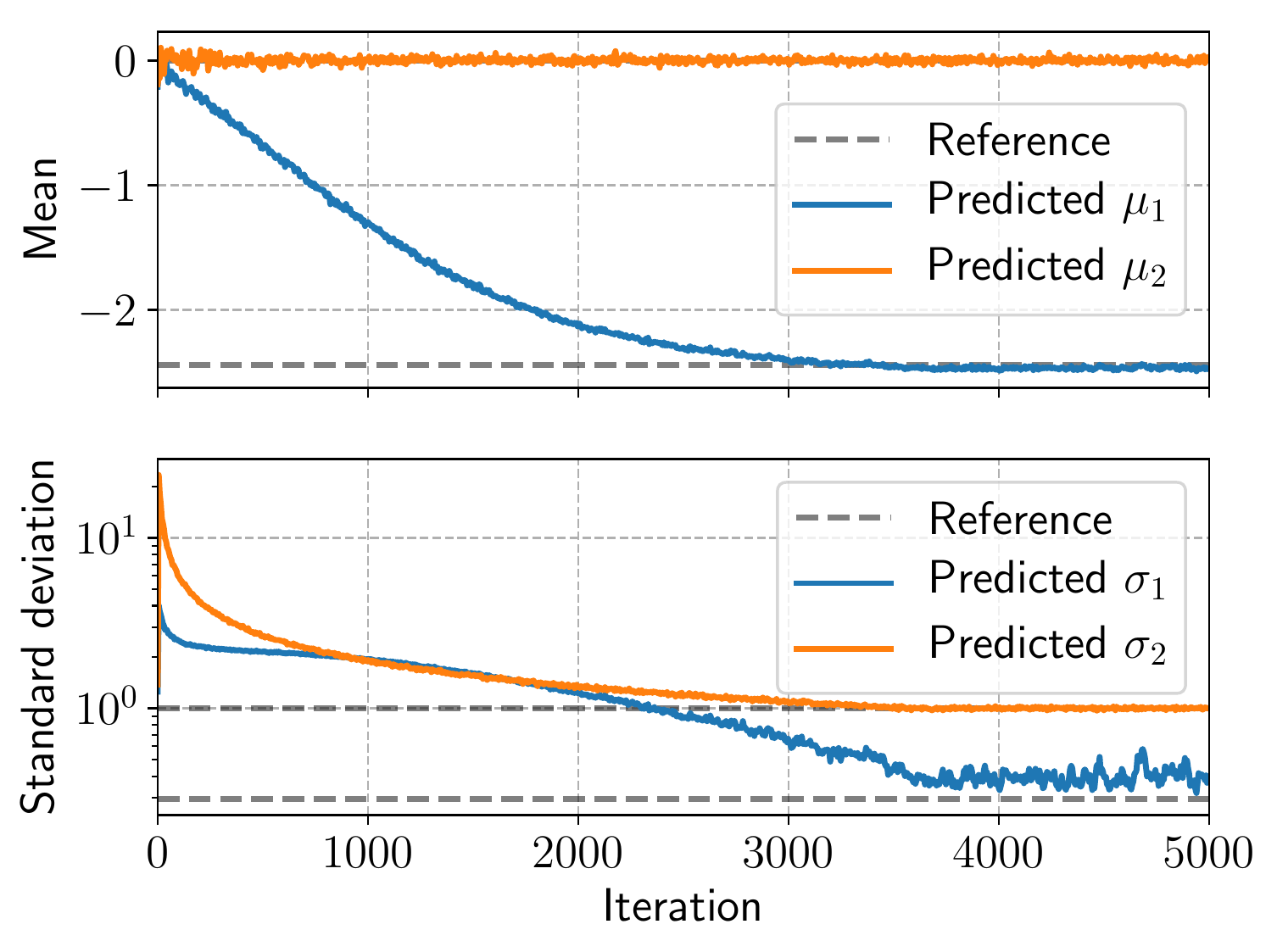}
    \label{fig:ch5_double_well_statistics}}
    \caption{Convergence of the KL divergences (left) and predicted statistics compared with reference estimates (right).}
    \label{fig:ch5_double_well_convergence}
\end{figure}

\figref{fig:ch5_double_well_kl_bounds} shows the overall convergence of the model, expressed in the form of the reverse KL divergence (\eqqref{eqn:ch5_method_reverse_kl}) and the forward KL divergence (\eqqref{eqn:ch5_method_forward_kl}); the latter, which relies on the data, is only used for illustrative purposes. Data for evaluating $D_\text{KL}\left( p\indtarg(\bx) \| q\indtheta(\bx) \right)$ were not used in the training process. We compare reference statistics (again based on data which were not used during training) with statistics estimated based on the efficient predictive distribution $q\indtheta(\bx)$ in \figref{fig:ch5_double_well_statistics}

\subsubsection{Predictive collective variables}
\label{sec:ch5_double_well_cv}

The proposed approach provides an efficient CG model that can be employed for predictive purposes, as described in the previous section. We claim that in addition to obtaining a CG model, we can provide relevant insights by identifying CVs of the system. In the double well example, one would expect the CV to be the $x_1$ coordinate that separates the two modes, where conformational changes are implied by moving along $x_1$.

\begin{figure}
    \centering
    \subfigure[$\beta \approx 0$]{\includegraphics[width=0.45\textwidth]{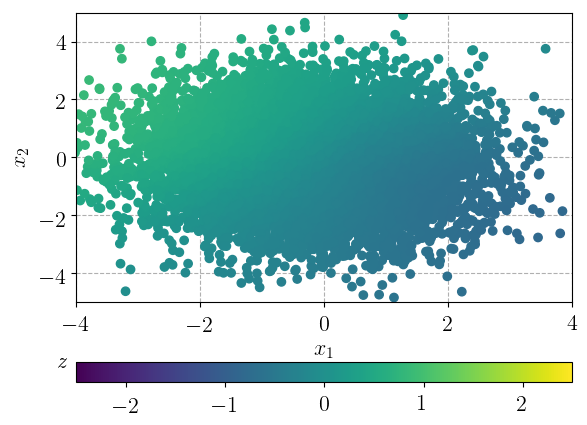}}
    \qquad
    \subfigure[$\beta \approx 0.36$]{\includegraphics[width=0.45\textwidth]{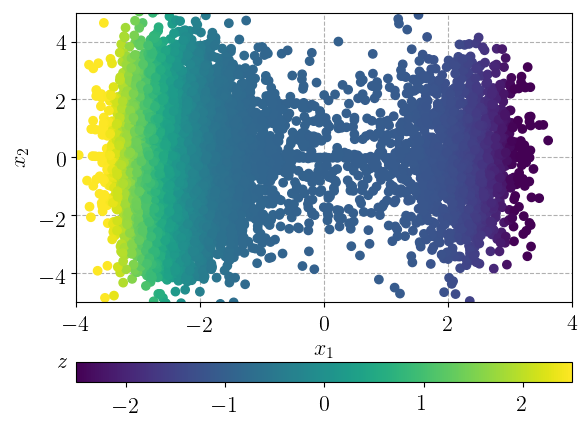}}
    \qquad
    \subfigure[$\beta \approx 0.7$]{\includegraphics[width=0.45\textwidth]{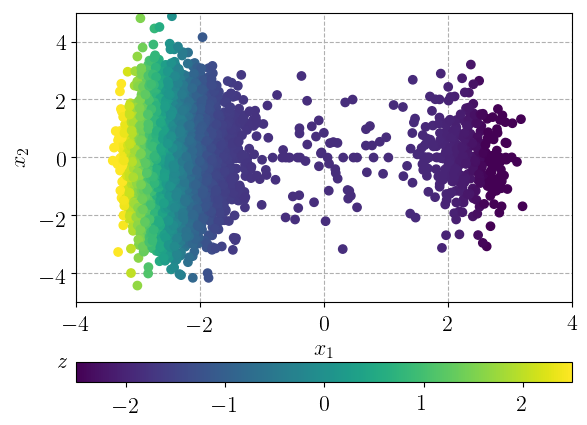}}
    \qquad
    \subfigure[$\beta \approx 1$]{\includegraphics[width=0.45\textwidth]{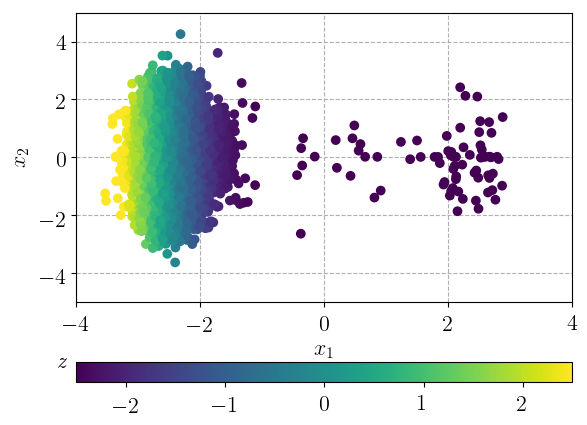}}
    \qquad
    \caption{Samples $\bxi \sim q\indtheta(\bx)$ at the indicated temperature $\beta$ are depicted as dots, whereas the assigned color of $\bxi$ corresponds to its latent CV obtained by the mean of $r\indphi(\bz|\bxi)$. The color bar below the images shows the color corresponding to the assigned value of the CV $\bz$ given $\bx$. The figure is based on $N=\num{1e4}$ samples $\bxi$.}
    \label{fig:ch5_double_well_cvs}
\end{figure}

To visualize the assigned CVs given samples $\bxi \sim q\indtheta(\bx)$, we plot samples as dots in \figref{fig:ch5_double_well_cvs}, while the color of the $\bxi$ is assigned based on the corresponding value of the CV. We note that for every $\bxi$ there exists a whole distribution of CVs $r\indphi(z|\bxi)$, as we work in a probabilistic framework. The assigned color in \figref{fig:ch5_double_well_cvs} is based on the mean of  $r\indphi(z|\bxi)$, which is obtained by evaluating $\bmu\indphi(\bxi)$.

The (color) gradient of $z$ with respect to $\bx$ is almost exactly parallel to the $x_1$-direction,  which implies that the revealed CV $z$ is (probabilistically) parallel to the $x_1$ axis and thus meets our expectations. The proposed approach reveals the relevant, slow, CV $x_1$ solely by evaluating $U(\bx)$ under $q\indtheta(\bx)$.

\clearpage

\subsection{ALA-2}
\label{sec:ch5_nummerical_illustrations_ala2}

\begin{figure}
	\centering
	\subfigure[Dihedral angels for ALA-2.]{
	\label{fig:ch5_numill_ala2_dihedral_angles}
	\includegraphics[width=0.4\textwidth]{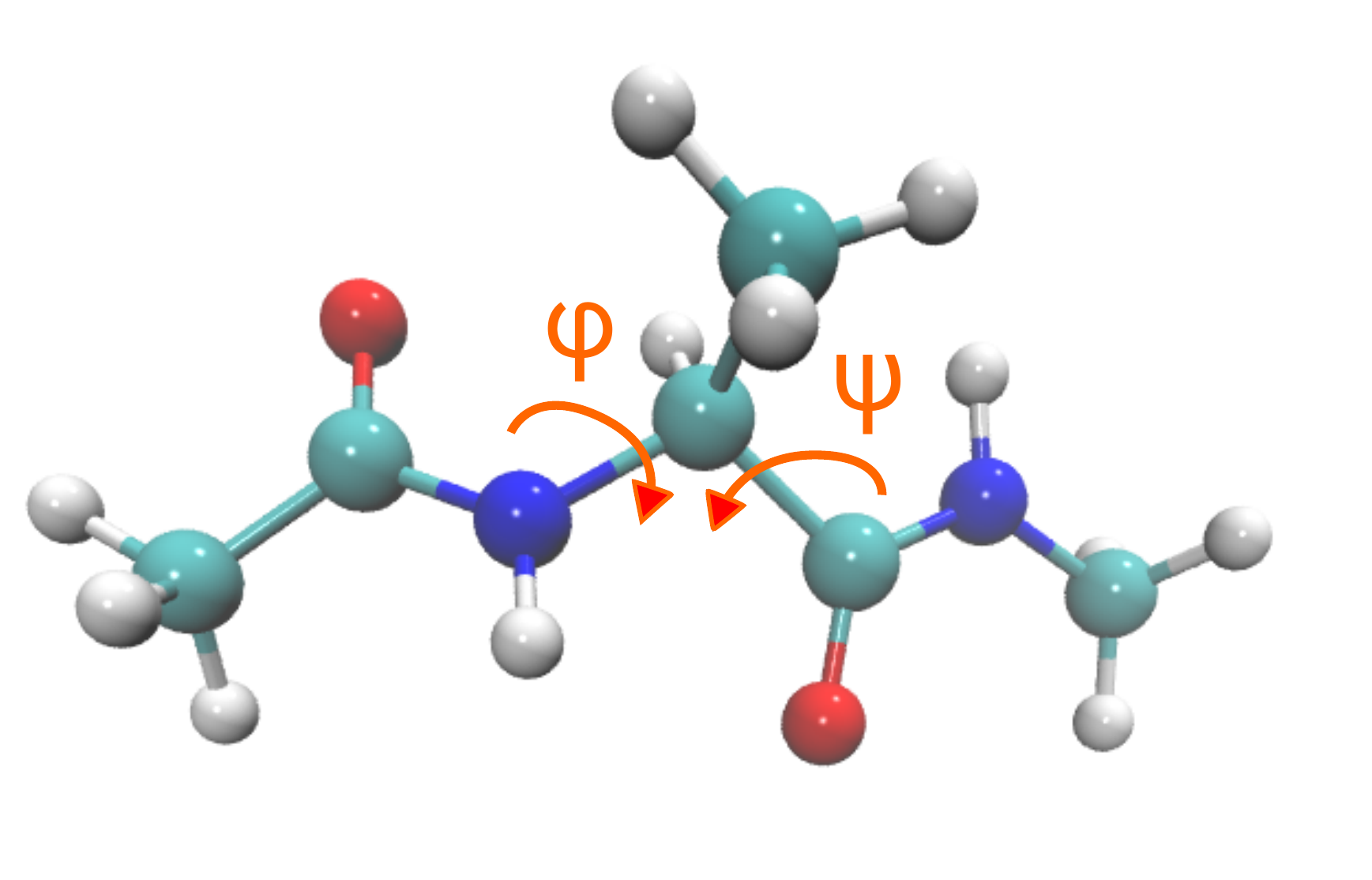}}
	\qquad
	\subfigure[Characteristic conformations according \cite{vargas2002}.]{
	\label{fig:ch5_numill_ala2_modes}
	\includegraphics[width=0.4\textwidth]{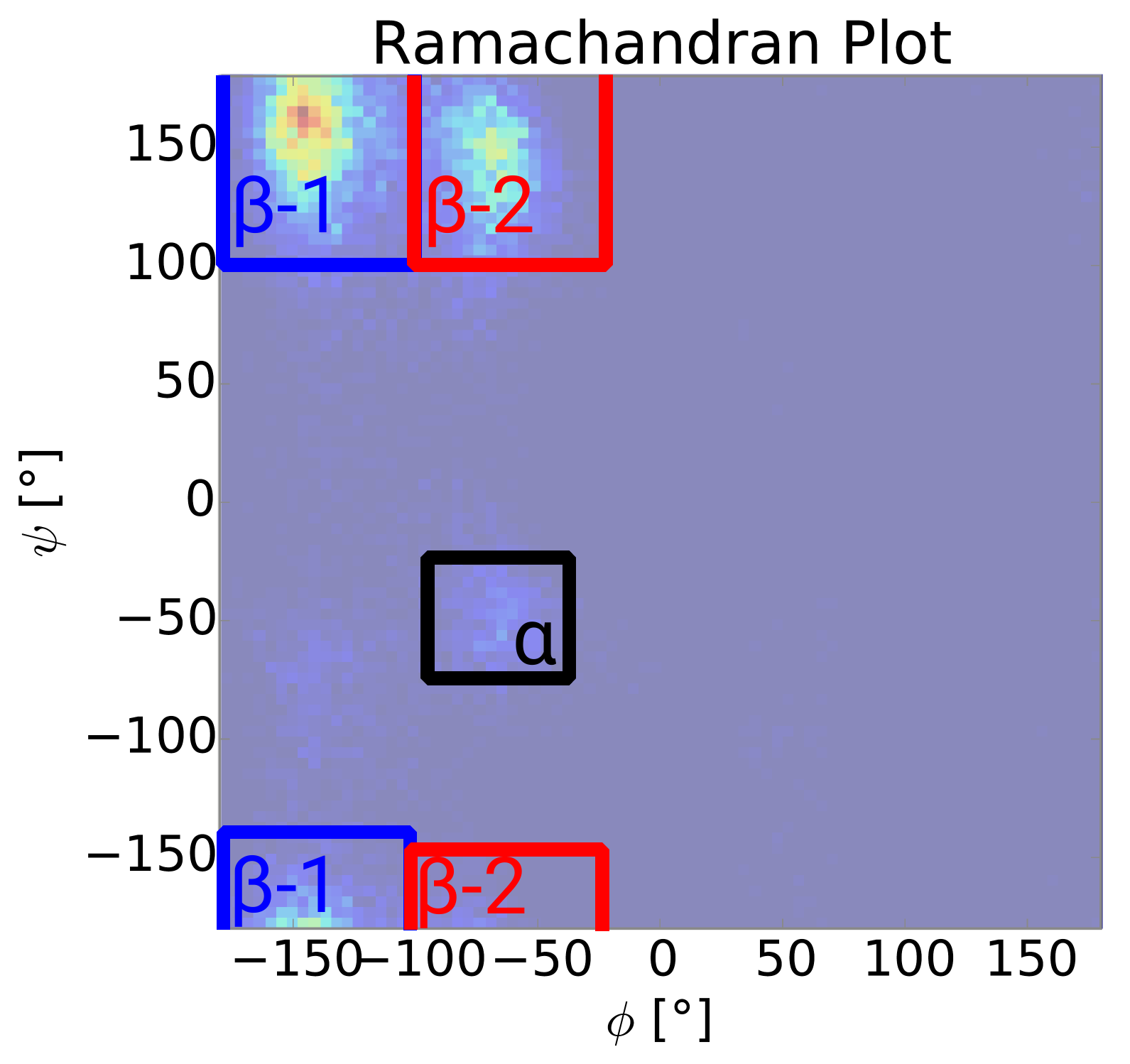}}
	\caption{Dihedral angles (left) and $(\phi, \psi)$ statistics of a reference simulation with characteristic modes (right).}
	\label{fig:ch5_numill_ala2_ref}
\end{figure}

After demonstrating the functionality of the proposed scheme for a double well potential energy, we are interested in addressing an atomistic system. The following is devoted to the CV discovery of alanine dipeptide (ALA-2) in the context of an implicit solvent. Characteristic coordinates of the ALA-2 peptide include the dihedral angles $(\phi, \psi)$, as shown in \figref{fig:ch5_numill_ala2_dihedral_angles}. Distinct combinations of the dihedral angles characterize three distinguishable ($\alpha$, $\beta\textnormal{-}1$, $\beta\textnormal{-}2$) conformations, as provided in the Ramachandran plot \cite{ramachandran1963} in \figref{fig:ch5_numill_ala2_modes} \cite{vargas2002}. The peptide consisting of 22 atoms can be described by 60 effective degrees of freedom (rigid body motion removed); however, we store the complete Cartesian coordinate vector $\bx$ with $\dim(\bx) = 66$, where six degrees of freedom are fixed. The exact representation of ALA-2 in $\bx$ with coordinate bookkeeping is given in the Appendix \ref{sec:ch5_appendix_ala2_coordinate_representation}.

\subsubsection{Reference model setting}

Applying the proposed methodology does not require the production of any reference atomistic trajectories. However, we are interested in comparing our obtained predictions from the generative CG model to reference observables estimated by a reference MD simulation. We refer to Appendix \ref{sec:ch5_appendix_ala2_sim_details} for all necessary details obtaining the MD trajectory.
Nevertheless, for the sake of evaluating forces, we need to specify system properties such as the force field, which in this case is AMBER ff96 \cite{sorin2005, depaul2010, allen1989}.
We employ an implicit water model based on the generalized Born approach  \cite{onufriev2004, still1990}, which serves as a solvent. The temperature of interest is $T=\SI{330}{K}$.

\subsubsection{Model specification}

The general model structure introduced earlier in  \secref{sec:ch5_methods_model} is also employed in the context of the ALA-2 setting. We mostly rely on findings in \cite{schoeberl2019_pcvs}, where an identical system was explored on the basis of data-driven forward KL divergence minimization.
All required details for the model are specified in Tables \ref{tab:ch5_ala2_encoder_actfct} and \ref{tab:ch5_ala2_decoder_actfct} for the encoder ($r\indphi(\bz|\bx)$) and decoder ($q\indtheta(\bx|\bz)$), respectively.
\begin{table}[]
	\centering
	\begin{tabular}{|p{2cm} p{2.5cm} p{2cm} p{2cm} p{2.5cm} |}
		\hline
		Linear layer & Input \linebreak dimension & Output dimension & Activation layer & Activation function \\
		\hline
		\hline
		$l\indphi^{(1)}$ & $\dim(\bx)=66$ & $d_1$ &  $a^{(1)}$ & SeLu\footnote{SeLu: $a(x) = 
			\begin{cases}
			\alpha (e^x -1) & \text{if } x<0\\
			x            & \text{otherwise}.
			\end{cases} $ See \cite{klambauer2017} for further details.}\\
		$l\indphi^{(2)}$ & $d_1$ & $d_2$ &  $a^{(2)}$ & SeLu \\
		$l\indphi^{(3)}$ & $d_2$ & $d_3$ &  $a^{(3)}$ & Log Sigmoid \footnote{Log Sigmoid: $a(x) = \log \frac{ 1 }{ 1 + e^{-x}} $}  \\
		$l\indphi^{(4)}$ & $d_3$ & $\dim(\bz)$ &  None & - \\
		$l\indphi^{(5)}$ & $d_3$ & $\dim(\bz)$ &  None & - \\
		\hline
	\end{tabular}
	\caption{Network specification of the encoding neural network with $d_{\{1,2,3\}}=170$.}
	\label{tab:ch5_ala2_encoder_actfct}
\end{table}
\begin{table}[]
	\centering
	\begin{tabular}{|p{2cm} p{2.5cm} p{2cm} p{2cm} p{2cm} |}
		\hline
		Linear layer & Input \linebreak dimension & Output dimension & Activation layer & Activation function \\
		\hline
		\hline
		$l\indtheta^{(1)}$ & $\dim(\bz)=2$ & $d_3$ &  $\tilde{a}^{(1)}$ & Tanh \\
		$l\indtheta^{(2)}$ & $d_3$ & $d_2$ &  $\tilde{a}^{(2)}$ & Tanh \\
		$l\indtheta^{(3)}$ & $d_2$ & $d_1$ &  $\tilde{a}^{(3)}$ & Tanh \\
		$l\indtheta^{(4)}$ & $d_1$ & $\dim(\bx)$ &  None & - \\
		\hline
	\end{tabular}
	\caption{Network specification of the decoding neural network with $d_{\{1,2,3\}} = 120$.}
	\label{tab:ch5_ala2_decoder_actfct}
\end{table}
Similar to the previous example in \secref{sec:ch5_nummerical_illustrations_double_well}, we employ a tempering scheme as introduced in \secref{sec:ch5_methods_training} and specified in Algorithm \ref{alg:ch5_method_kl_tempering} with initial $\beta_0 = \num{1e-14} \cdot \beta_K$, while the target temperature  is defined by $\beta_K = \frac{1}{k_B T}$ and $T= \SI{330}{K}$.
The inverse temperature $\beta_0$ occurs as a multiplicative factor, multiplying the potential energy $U(\bx)$. For gradient estimation, the interatomic force $\mathbf{F}(\bx)$ is multiplied by $\beta_k$. In the variational approach presented in this work, we evaluate the force field under samples from $q\indtheta(\bx)$. However, when $q\indtheta(\bx)$ has not yet learned, samples $\bxi$ will potentially yield high-energy states associated with large forces. According to experimental results, the magnitude of $\mathbf{F}(\bx)$ in early training stages reaches $\num{\pm 1e-18}$. Therefore, the initial inverse temperature is chosen such that $\beta_0 \mathbf{F}(\bx)$ evaluated under $q\indtheta(\bx)$ yields values of $\pm \num{1e1}$. This implies that the embedded physics, expressed by interatomic forces $\mathbf{F}(\bx)$, are weak in the early training stages and are emphasized as the learning process proceeds with increasing $\beta_k$.
For further details, refer to Appendix \ref{sec:ch5_appendix_method_grad_norm}.

The stochastic optimization algorithm is ADAM \cite{kingma2015_adam} with $\alpha = 0.001,\beta_1= 0.9, \beta_2 =  0.999, \epsilon_{\text{ADAM}}=\num{1.0e-8}$. We employ $J=\num{2000}$ samples for computing expectations with respect to $q\indtheta(\bx)$ throughout the training process. Initially, in the early training stages, using fewer samples does not influence the training. The number of samples should be increased once the model has been refined and comes closer to $p\indtarg(\bx)$.

\subsubsection{Collective variables}

\begin{figure}
    \centering
    \subfigure[~Iteration 700.]{\includegraphics[width=0.4\textwidth]{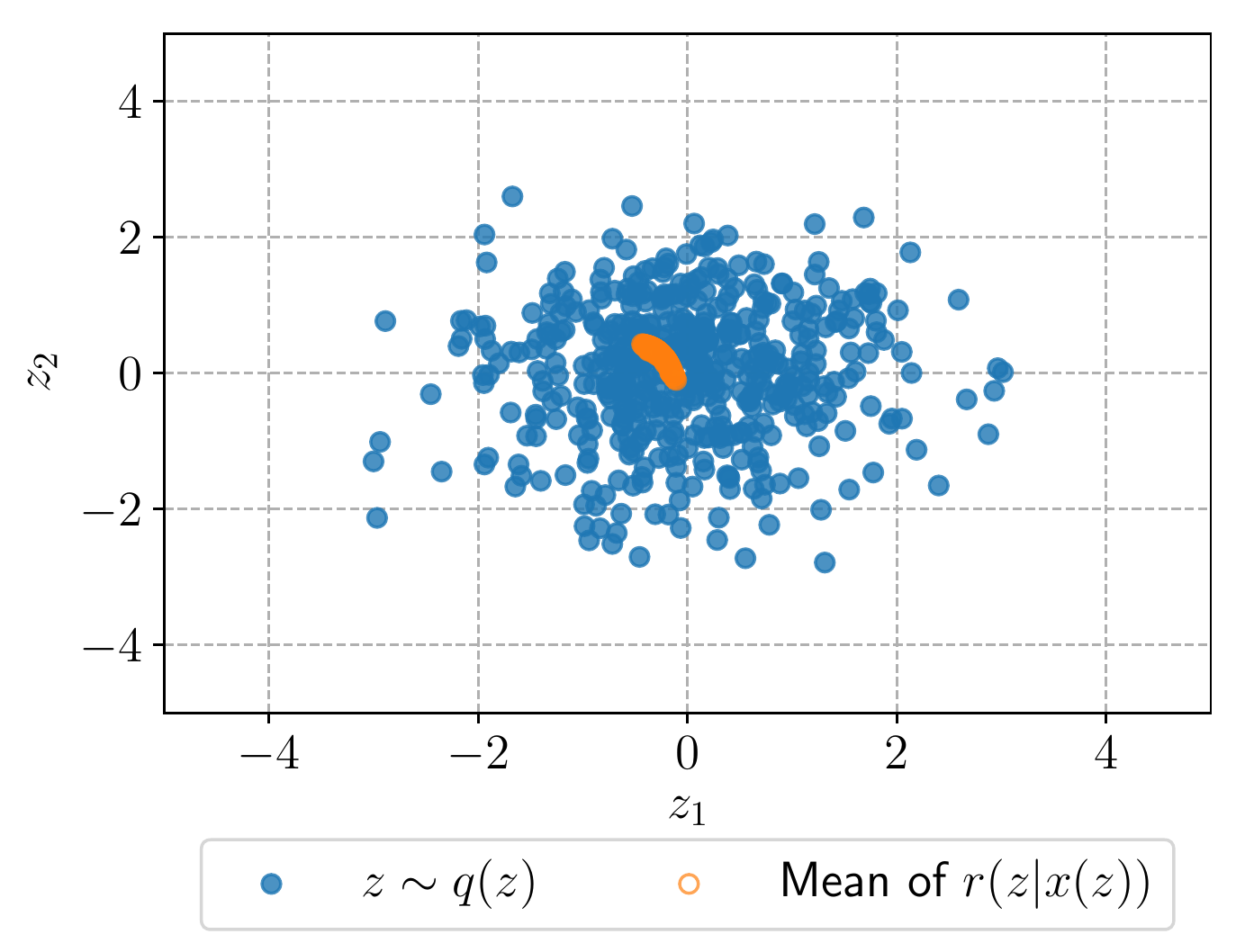}}
    \subfigure[~Iteration 2200.]{\includegraphics[width=0.4\textwidth]{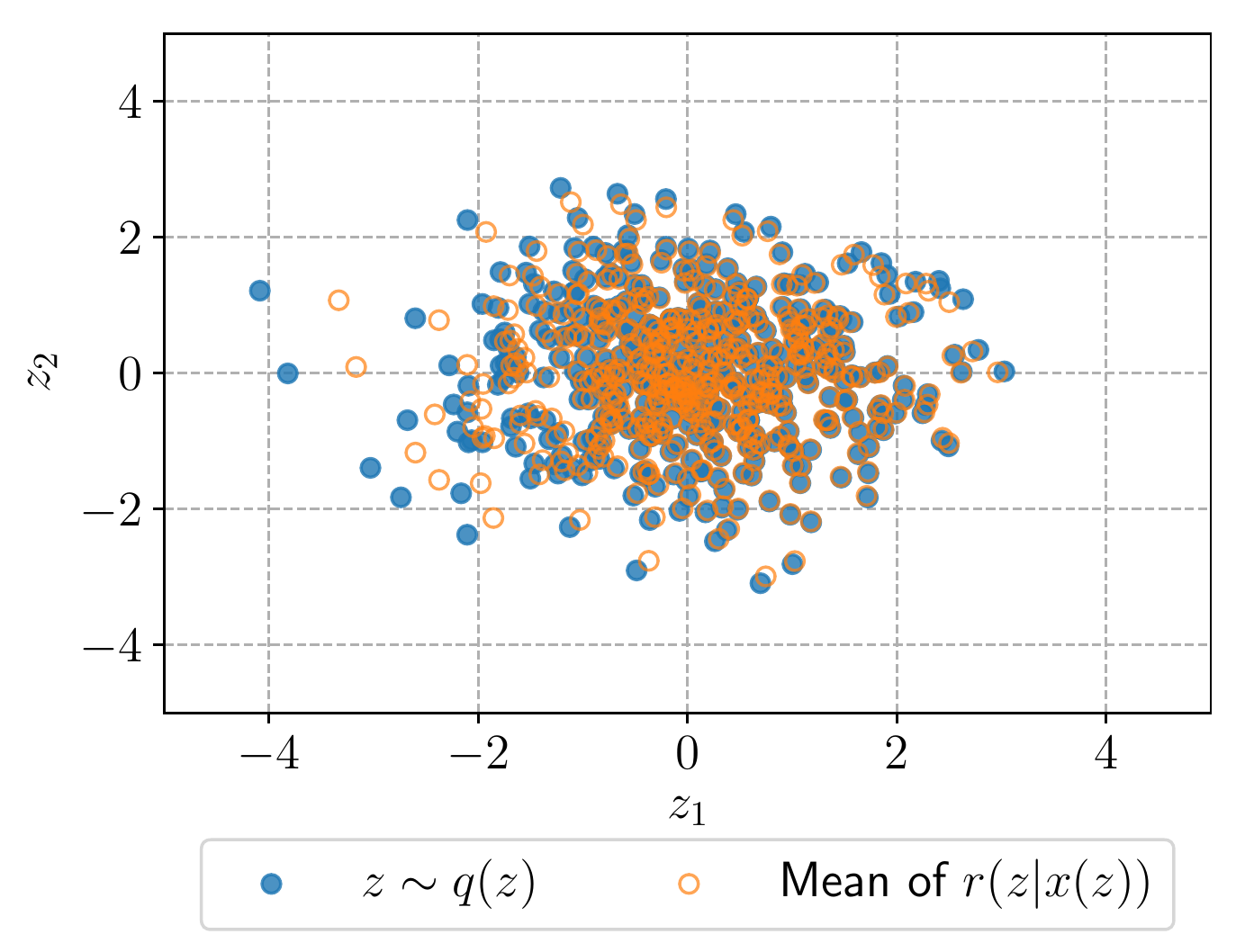}}
    \subfigure[~Iteration 7200.]{\includegraphics[width=0.4\textwidth]{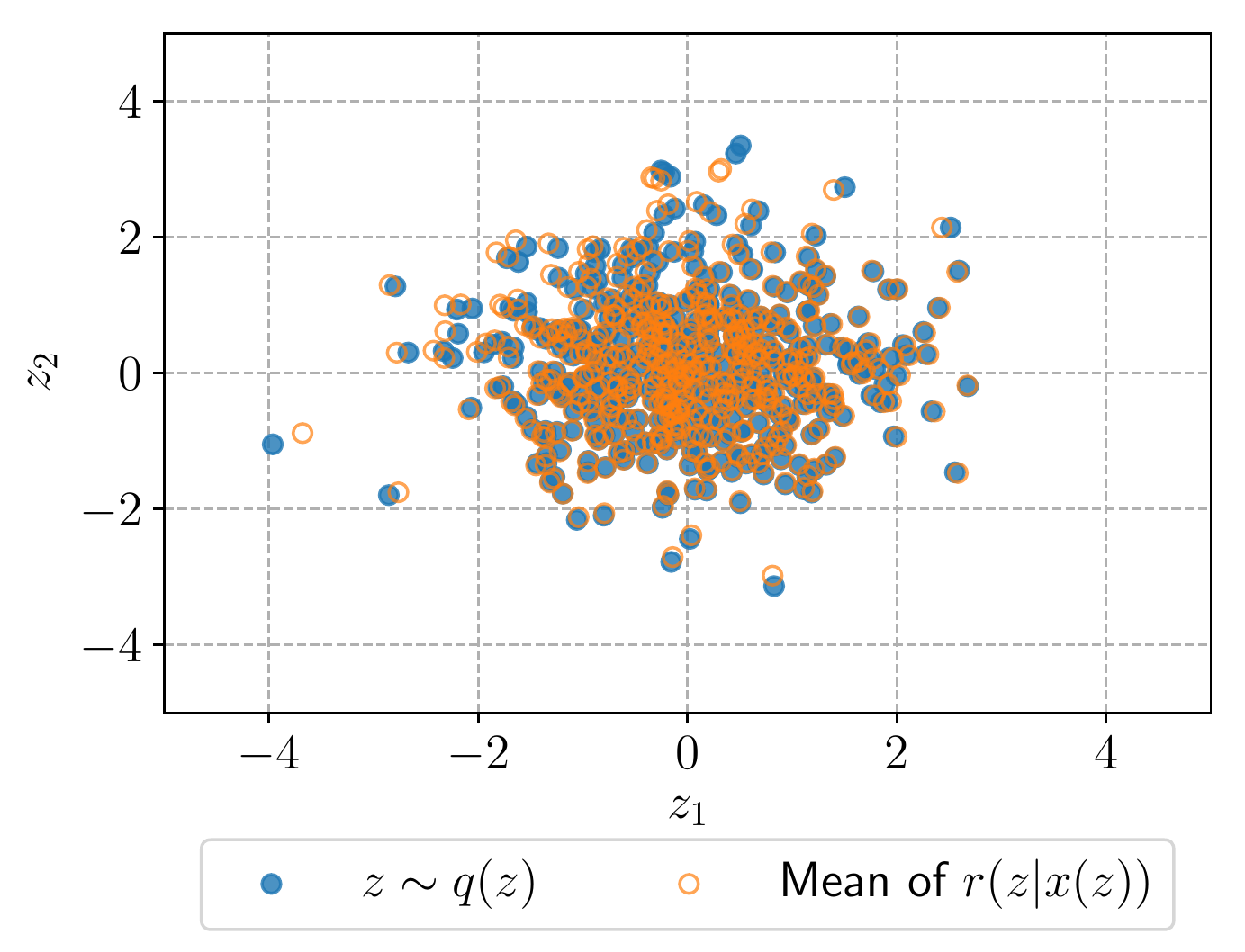}}
    \caption{Samples from $q\indtheta(\bz)$ (blue, filled) are decoded with $q\indtheta(\bx|\bz)$ and encoded with $r\indphi(\bz|\bx)$ (orange, no facecolor). We consider the means of aforementioned distributions for performing the decoding and encoding processes. Re-encoding the decoded $\bzi$ matches its origin.}
    \label{fig:ch5_ala2_latent_de_encoding_training}
\end{figure}

When training the model with its  encoder and decoder components $r\indphi(\bz|\bx)$ and $q\indtheta(\bx|\bz)$, it is important that these consistently map a generated sample $\bzi \sim q(\bz)$ to $\bxi$ through the decoder $q\indtheta(\bx|\bz)$, and from the decoded atomistic configuration  $\bxi$ back to its origin, the value of the CV $\bzi$ it has been generated from. The projection from $\bxi$ to the according CV is enabled through the encoder  $r\indphi(\bz|\bx)$. After some initial iterations optimizing $(\bphi, \btheta)$, the encoder and decoder work consistently as depicted in \figref{fig:ch5_ala2_latent_de_encoding_training}.

\begin{figure}
	\centering
	\includegraphics[width=0.6\textwidth]{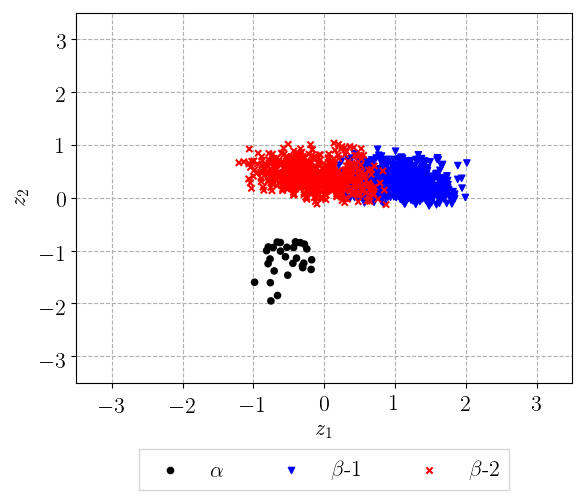}
	\qquad
	\caption{Representation of $\bz$-coordinates of test data assigned by the mean of $r\indphi(\bz|\bz)$, which we learn by minimizing the reverse KL divergence without reference data.
	Characteristic conformations of ALA-2 are indicated in: $\alpha$ black, $\beta\textnormal{-}1$ blue, and $\beta\textnormal{-}2$ red color. Without any prior physical information and in the complete absence of any data, the encoder identifies physically relevant coordinates, which are related to $\phi, \psi$ values.
	}
	\label{fig:ch5_ala2_lat_rep}
\end{figure}

In \figref{fig:ch5_ala2_lat_rep} we utilize the identified encoder $r\indphi(\bz|\bx)$, which assigns CVs to an input atomistic configuration, for encoding a reference test dataset. This dataset has not been used for training and is used here solely for visualization purposes. The test data (generated according to Appendix \ref{sec:ch5_appendix_ala2_sim_details}) contains atomistic configurations from multiple characteristic modes based on their dihedral angle values $(\phi, \psi)$ as shown in \figref{fig:ch5_numill_ala2_modes}.
Given a datum from the test dataset $\bxi$, we can assign the corresponding value of its CV by employing the mean $\bmu\indphi(\bxi)$ of the approximate posterior over the latent variables $r\indphi(\bz|\bx)$. The assigned CV depicts the pre-image of the atomistic configuration in the reduced CV space.
It is important to note in \figref{fig:ch5_ala2_lat_rep} that atomistic configurations belonging to characteristic conformations  ($\alpha, \beta\textnormal{-}1, \beta\textnormal{-}2$) are identified by $r\indphi(\bz|\bx)$ and form clusters in the CV space.
We note that the conformations $\beta\textnormal{-}1, \beta\textnormal{-}2$ interleave with each other in regions around $z_1=0$. An explanation for this overlap is the similarity of $(\phi, \psi)$ combinations in the Ramachandran plot in  \figref{fig:ch5_numill_ala2_modes}. Separate from the $\beta$ configurations is the cluster associated with $\alpha$ configurations in the CV space. The latter differ significantly with respect to the $(\phi, \psi)$ pairs from the $\beta$ conformations.

The implied similarity of, e.g., $\beta$ conformations in the CV space is in accordance with the expectations on dimensionality reduction methods. Similar atomistic---or, in general, observed---coordinates should map to similar regions in their latent lower-dimensional embedding, as emphasized in \cite{rohrdanzclementi2013}. This is achieved in, e.g.,  multidimensional scaling \cite{troyer1995} or isomap \cite{tenenbaum2000}. The presented dimensionality reduction relies solely on evaluating the force field $\mathbf{F}(\bx)$ at generated samples from $q\indtheta(\bx)$, without using any data, and is differentiable with respect to $\bx$.

\begin{figure}
	\centering
	\includegraphics[width=0.7\textwidth]{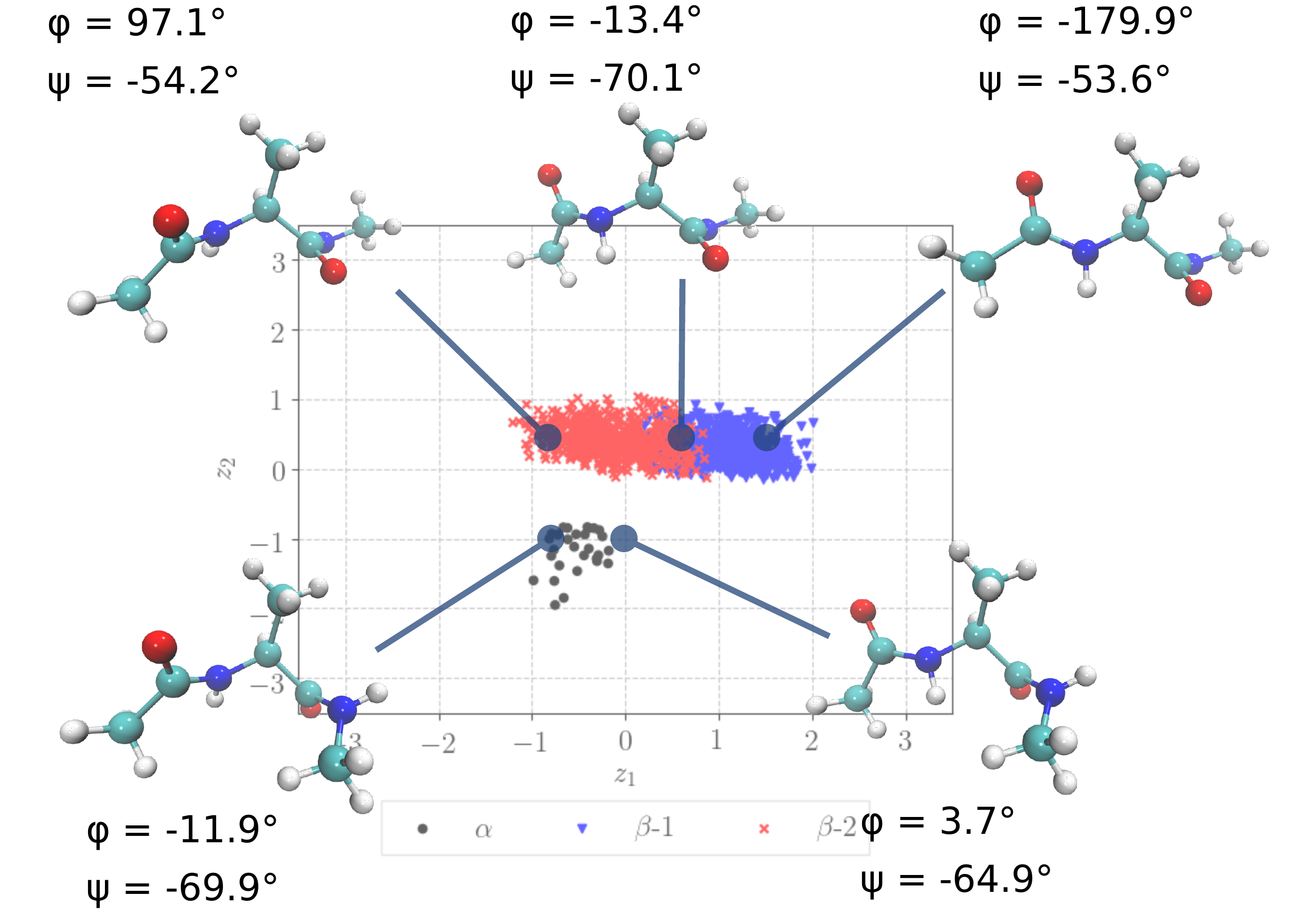}
	\caption{\emph{Predicted} configurations $\bx$ (including dihedral angle values) with $\bmu\indtheta(\bz)$ of $p\indtheta(\bx|\bz)$. As one moves along the $z_1$ axis, we obtain for the given CVs atomistic configurations $\bx$ reflecting the conformations $\alpha$, $\beta\textnormal{-}1$, and $\beta\textnormal{-}2$.
	All rendered atomistic representations in this work were created by VMD~\cite{humphrey1996_vmd}.}
	\label{fig:ch5_ala2_lat_rep_prediction}
\end{figure}

The hidden and lower-dimensional physically characteristic generative process is emphasized further in \figref{fig:ch5_ala2_lat_rep_prediction}. We illustrate \emph{predicted} atomistic configurations $\bx$ given the marked (circle) values of the CVs $\bz$. The change of characteristic $(\phi,\psi)$ dihedrals can be observed by moving from the red ($\beta\textnormal{-}2$) to the blue region ($\beta\textnormal{-}1$) in the CV space and observing the configurational change in the predicted atomistic configurations, given the indicated CVs. The depicted atomistic configurations are obtained using the input CV $\bz$ and the mean of $q\indtheta(\bx|\bz)$, which is expressed as a neural network with $\bmu\indtheta(\bz)$. The probabilistic decoder  $q\indtheta(\bx|\bz)$ is a distribution, implying that given one value of the CV, several atomistic realizations can be produced. For illustrative reasons we represent the mean $\bmu\indtheta(\bz)$.

\begin{figure}
	\centering
	\includegraphics[width=0.6\textwidth]{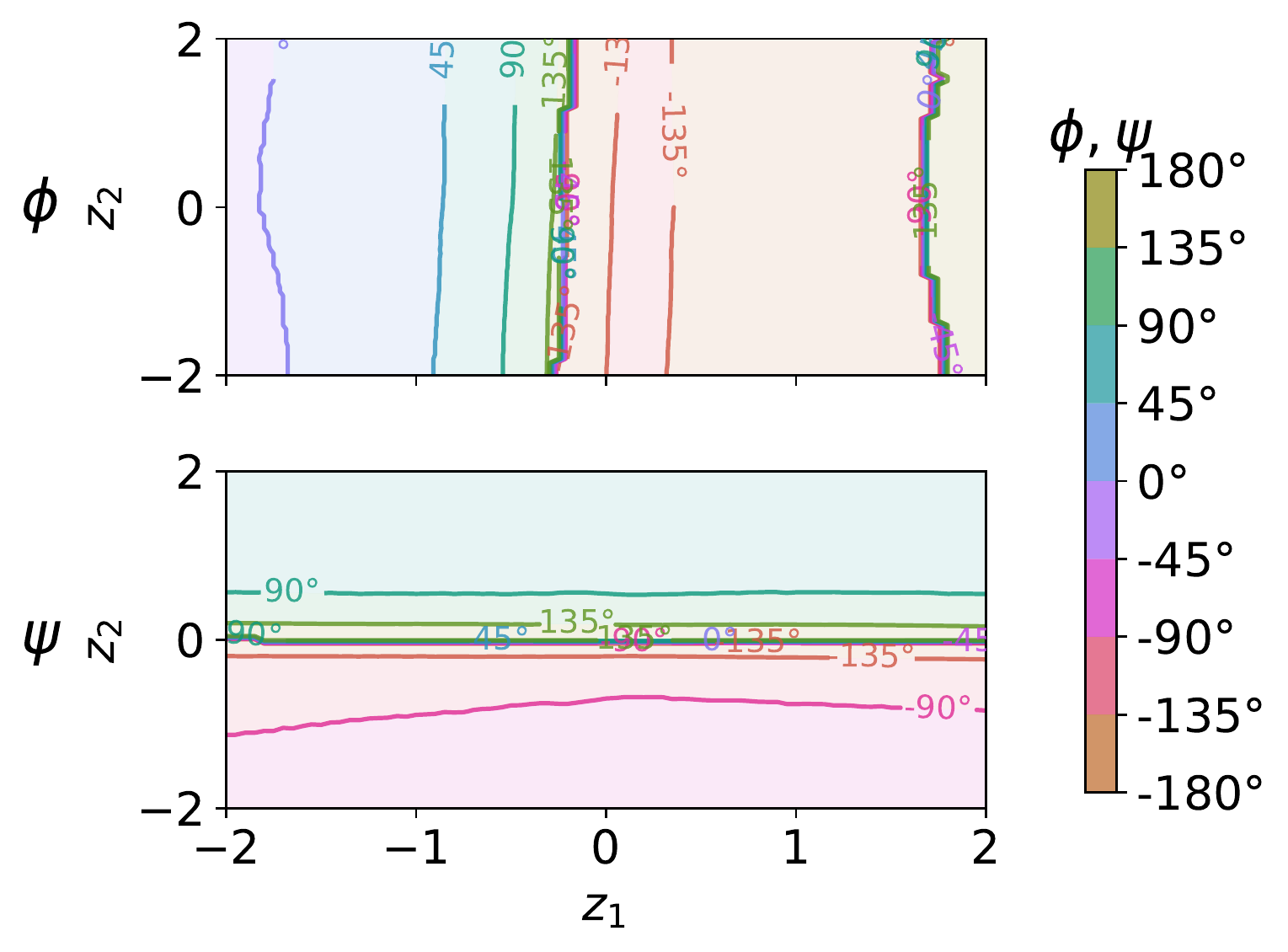}
	\caption{Predicted dihedrals $(\phi, \psi)$ for the latent CVs. The depicted  $(\phi, \psi)$ values were obtained from atomistic configurations given a CV value $\bz$ through the mean of $q\indtheta(\bx|\bz)$, $\bmu\indtheta(\bz)$.}
	\label{fig:ch5_ala2_phi_psi}
\end{figure}

To obtain a better understanding of the meaning of identified CVs in terms of the dihedral angles $(\phi, \psi)$, we visualize them by mapping given values of $\bz$ to atomistic configurations and compute the $(\phi,\psi)$ values assigned to the corresponding $\bz$, as shown in  \figref{fig:ch5_ala2_phi_psi}.
Again, in the probabilistic model, we draw multiple atomistic realizations $\bx$ given one CG representation $\bz$. The realization for a given $\bz$ fluctuates in terms of bonded vibrations rather than any change in the dihedrals $(\phi, \psi)$. We observe a strong correlation between  $(\phi,\psi)$  and the CVs $\bz$.

\begin{figure}
    \centering
    \includegraphics[width=\textwidth]{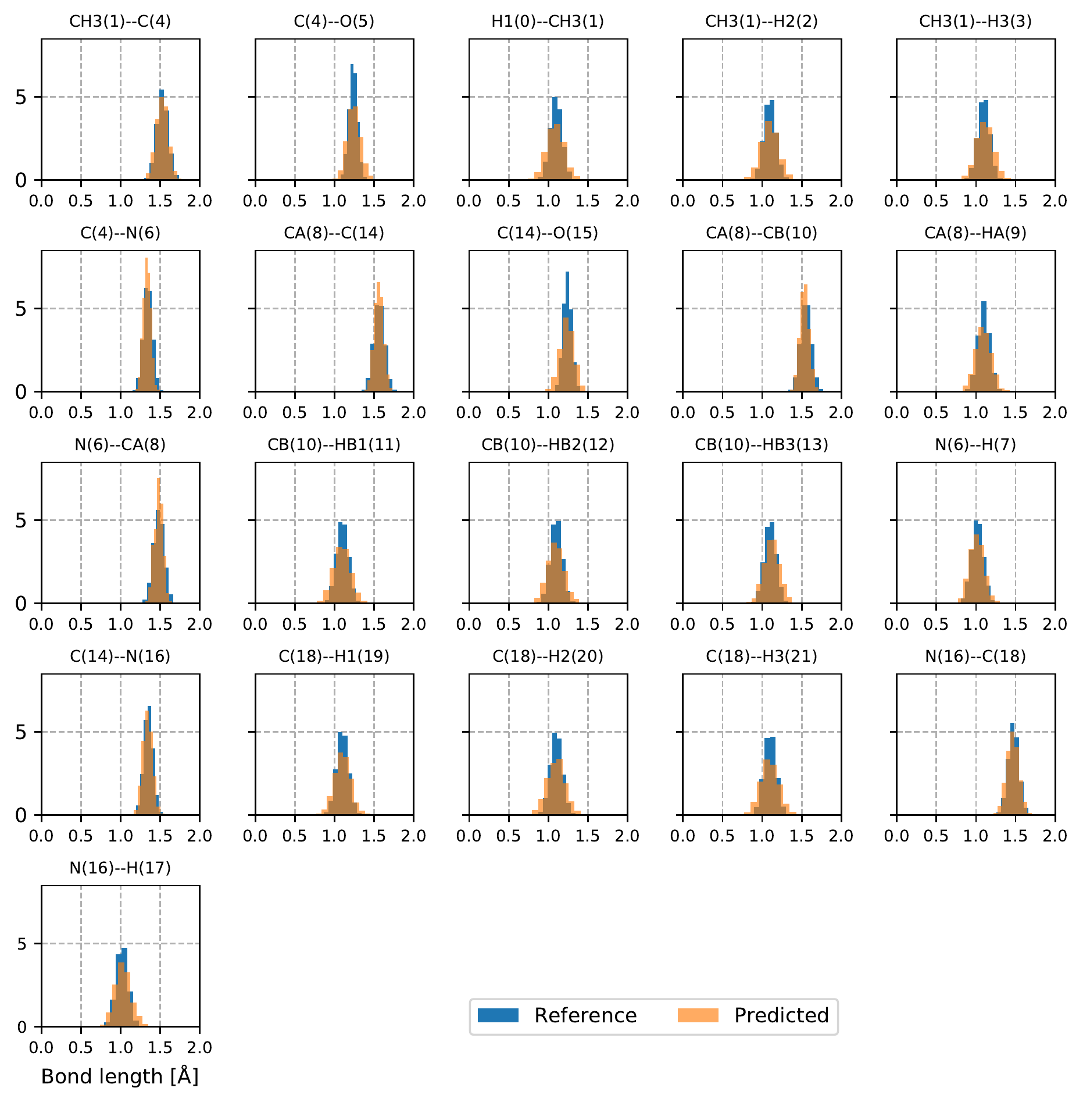}
    \caption{Bonding distance statistics. In ALA-2, bonded atoms of a reference simulation (blue) compared with histograms of the bond lengths of the \emph{predicted} atomistic ensemble based on $q\indtheta(\bx)$ (semi-transparent in the foreground in orange). The titles of the subplots indicate the relevant atom names, and the corresponding atom id of the structure file of ALA-2 as provided in Appendix  \ref{sec:ch5_appendix_ala2_coordinate_representation} is shown in brackets. The physics, in the form of bonding distances, is well maintained in the generated realizations. Predictive estimates are obtained by employing $J=\num{2000}$ samples of $q\indtheta(\bx)$, and the reference is based on $N=\num{4000}$ MD snapshots.}
    \label{fig:ch5_ala2_bond_lengths}
\end{figure}

\begin{figure}
    \centering
    \subfigure[]{\includegraphics[width=0.3\textwidth]{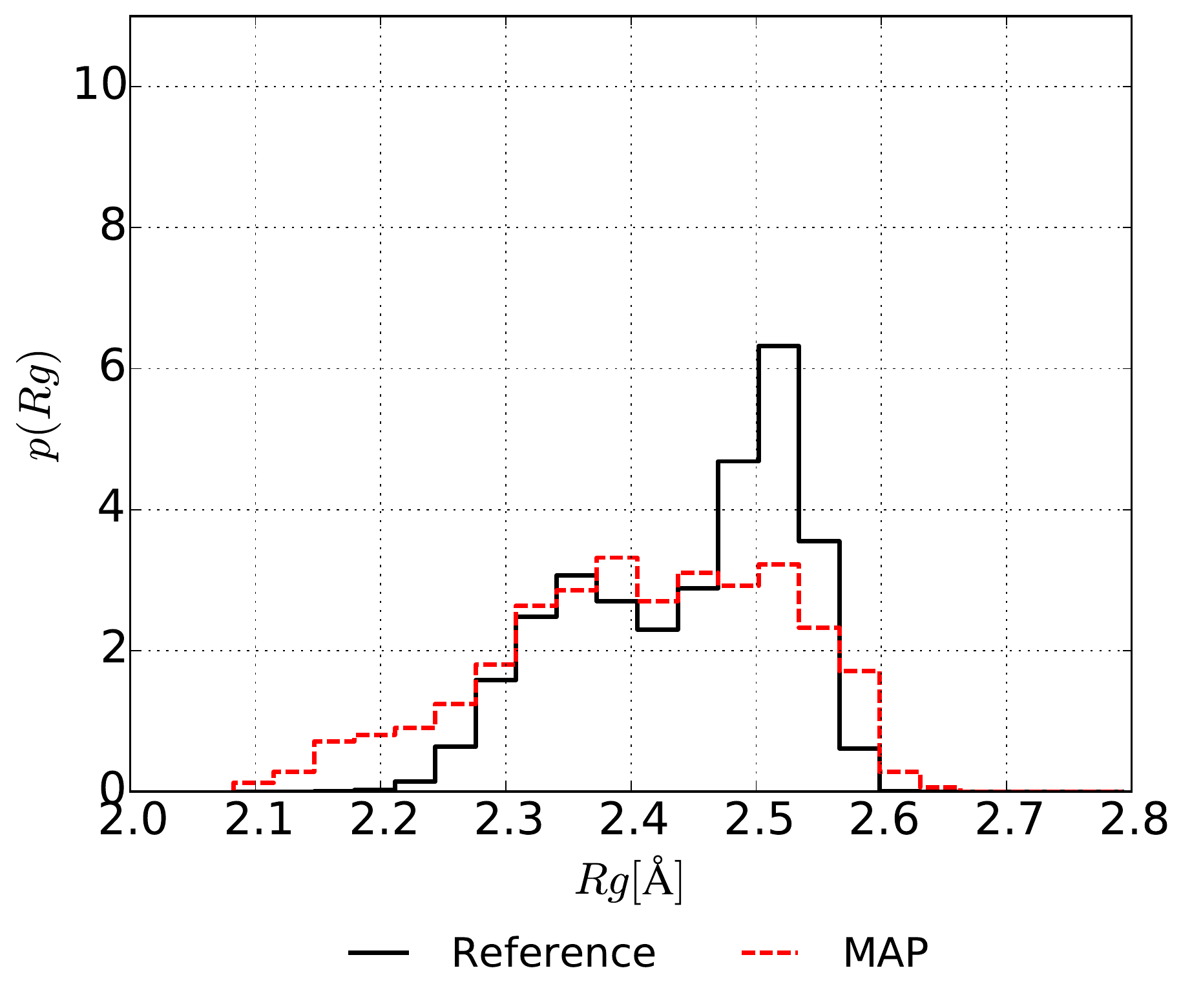}}
    \qquad
    \subfigure[]{\includegraphics[width=0.3\textwidth]{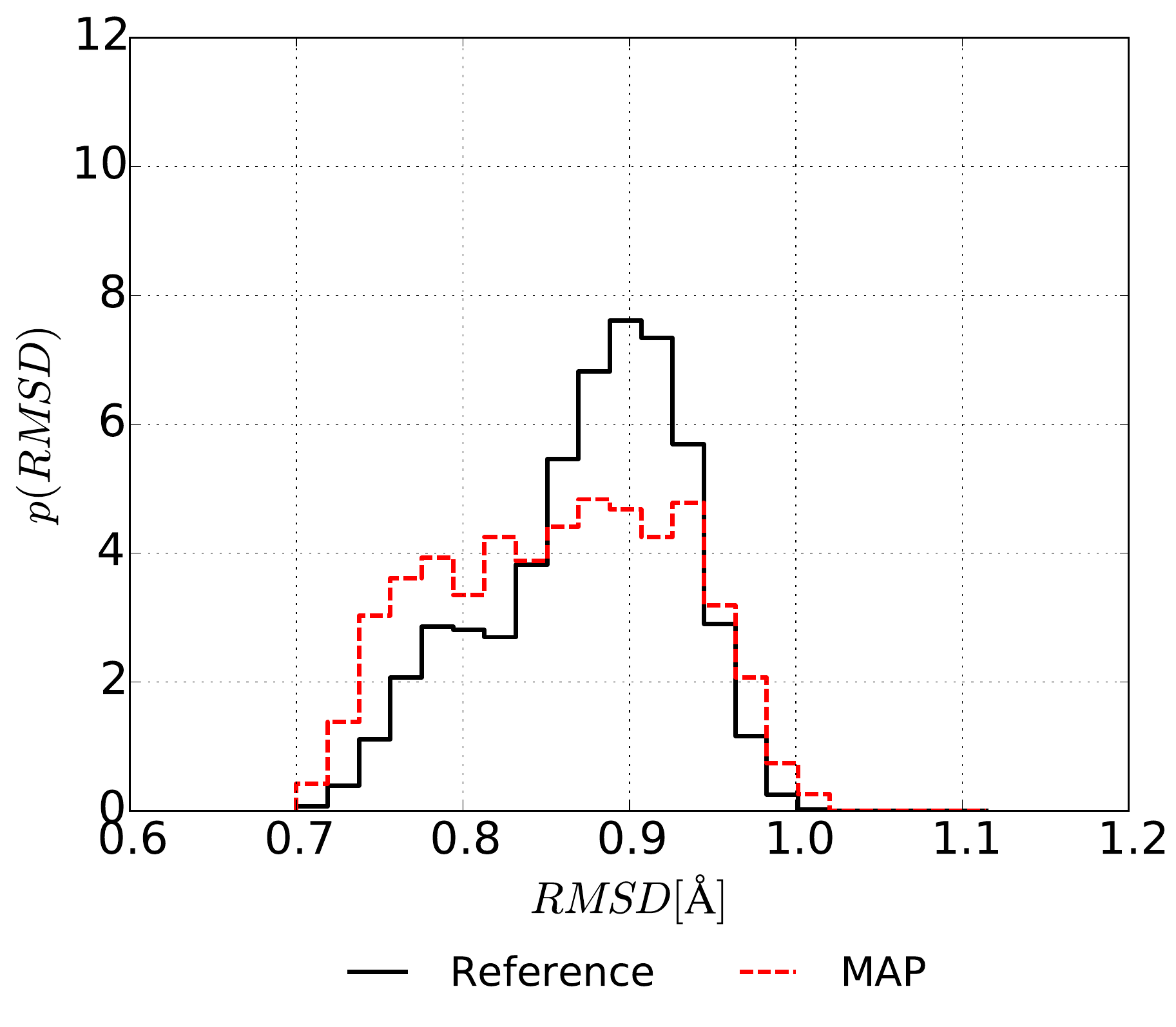}}
    \caption{Predicted observables compared with reference estimates. Radius of gyration (left) and root-mean-squared deviation (right). Predictive estimates are obtained by employing $J =\num{2000}$ samples of $q\indtheta(\bx)$ and the reference by $N=\num{4000}$ MD snapshots. Observables are estimated according to Appendix \ref{sec:ch5_appendix_ala2_observable_estimation}.}
    \label{fig:ch5_ala2_properties}
\end{figure}

In addition to the visual assessment given in \figref{fig:ch5_ala2_lat_rep_prediction}, we show quantitatively that the structural properties of atomistic configurations generated through $q\indtheta(\bx)$ truly capture those of a reference trajectory at $T=\SI{330}{K}$, as shown in \figref{fig:ch5_ala2_bond_lengths}.
\figref{fig:ch5_ala2_bond_lengths} provides histograms over bonding distances over all bonded atoms in the system. Reference statistics of bond lengths are compared with those based on generated samples of the predictive distribution $q\indtheta(\bx)$. 
\figref{fig:ch5_ala2_properties} provides estimated observables based on the predictive model and a reference trajectory. The observables are computed as explained in Appendix \ref{sec:ch5_appendix_ala2_observable_estimation}.

\section{Summary and outlook}
\label{sec:ch5_conclusions}

We have presented a variational approach for revealing CVs of atomistic systems that additionally yield a predictive CG model. We circumvent the need for reference data, which is supposed to provide an approximation of the target distribution $p\indtarg(\bx)$. The simulation of complex biochemical systems and thus the obtained data may suffer from bias owing to insufficiently exploring all relevant conformations. Conformations are separated by high free-energy barriers, which hamper efficient exploration with brute-force MD \cite{sittel2018_cv_nonlinear_map}.
The presented variational coarse-graining and CV discovery approach is guided by evaluating interatomic forces under the predictive distribution $q\indtheta(\bx)$, where sampling is computationally efficient. By embedding the atomistic force components, $q\indtheta(\bx)$ learns from the target distribution $p\indtarg(\bx)$. We derived an upper bound on the reverse KL divergence, in which all terms are tractable, and discuss the physical underpinning of the components involved. The derived upper bound is subject to minimization with respect to all model parameters. We provide a variance-reducing gradient estimator based on reparametrization. Whereas variational approaches are known for being mode focusing, remedy provides the introduced consistent tempering scheme, alleviating the simultaneous learning of modes.
We demonstrate the proposed algorithmic advances with a double well potential energy and the ALA-2 peptide. Characteristic CVs have been identified by the introduced optimization objective.

The following steps will be pursued in continuation of this work. Atomistic forces and thus gradients span many orders of magnitude at initial iterations. This could lead to numerical instabilities. Thus, we are interested in synthesizing the advantages of the forward and reverse KL divergences in the context of atomistic systems. We propose an adaptive learning scheme that may rely in its early training stages on a few data points. These are not required to reflect the whole phase space, but it is important to have a basis for learning, e.g., the structure of the atomistic system with its approximate bond lengths. This eases the problem of evaluating $\mathbf{F}(\bx)$ for the un-physical realizations that may be predicted by $q\indtheta(\bx)$ in an early training phase:
\begin{equation}
    \mathcal{E} =  \gamma D_\text{KL} \left( q\indtheta(\bx)   \| p\indtarg (\bx) \right) + (1-\gamma) D_\text{KL} \left( p\indtarg (\bx) \| q\indtheta(\bx) \right).
    \label{eqn:ch5_outlook_mixture_objective}
\end{equation}
With $\gamma \in [0, 1]$ weighting the overall contribution from the reverse and forward KL divergences, we use an adaptive weight,  $\gamma(k)$, which implies dependence on the current iteration $k$. With the proceeding learning process (increasing $k$), $\gamma$ could increase up to $\gamma=1$, so that we fully rely on the variational approach and thus the associated physics expressed by the potential $U(\bx)$ and forces $\mathbf{F}(\bx)$.
Minimizing the above objective \eqqref{eqn:ch5_outlook_mixture_objective} with respect to model distributions synthesizes the findings of this work and those of \cite{schoeberl2019_pcvs}.

We furthermore propose the employment of the obtained $q\indtheta(\bx)$ for predictive purposes of systems at different temperatures. This can be achieved by obtaining the implicitly learned predictive potential expressed in terms of fine-scale coordinates at $\beta\indtarg$ based on $q\indtheta(\bx)$:
\begin{equation}
    U\indtheta^{\text{pred}}(\bx) = - \frac{1}{\beta\indtarg} \log q\indtheta(\bx) + \mathrm{const.}
    \label{eqn:ch5_outlook_u_pred}
\end{equation}
Assuming we are interested in simulating the same system at $\beta_{\text{new}}$ where  $\beta_{\text{new}} \neq \beta\indtarg$, we can readily provide a generalized predictive distribution for any $\beta_{\text{new}}$,
\begin{equation}
    \Tilde{q}\indtheta(\bx) \propto e^{-\beta_{\text{new}} U\indtheta^{\text{pred}}(\bx) },
\end{equation}
by employing the predictive potential $U\indtheta^{\text{pred}}(\bx)$ defined in \eqqref{eqn:ch5_outlook_u_pred}.

Finally, we emphasize the most relevant findings of this work. We have reformulated the identification of CVs as an optimization problem, which additionally provides a predictive CG model. CVs are revealed in the absence of any prior physical knowledge or insight, and thus in the absence of any system-dependent assumptions. Instead of relying on reference data, we employ the minimization of the reverse KL divergence and develop an inference scheme in the context of atomistic systems. Thus, the optimization is solely guided by the \emph{evaluation} of the potential $U(\bx)$ and/or forces $\mathbf{F}(\bx)$ at samples of the predictive distributions $q\indtheta(\bx)$. We have also developed an adaptive tempering scheme based on findings of \cite{bilionis2012_free_energy}.

%% file: appendix.tex
\section{Relation with Expectation-Propagation}

This section emphasizes the relationship of hierarchical variational models with expectation-propagation (EP) \cite{minka2001}.

The following is not directly relevant to optimization of the objective Equation \eqref{eqn:ch5_method_reverse_kl}, but it shows the existence of an upper bound of the entropy term $-\mathbb{E}_{q(\bx)} \left[ \log  q(\bx) \right]$. Similar to Equation \eqref{eqn:ch5_methods_entropy_lowerbound_1} one denotes,
\begin{align}
    -\mathbb{E}_{ q(\bx)} \left[ \log  q(\bx) \right] &=  -\mathbb{E}_{ q(\bx)} \left[ \log  q(\bx) - D_{KL}\left( q(\bz|\bx)|| q(\bz|\bx) \right) \right] \nonumber \\ 
    &\leq \mathbb{E}_{ q(\bx)} \left[ -\log  q(\bx) + D_{KL}\left( r(\bz|\bx)|| q(\bz|\bx) \right) \right] \nonumber \\
    &= \mathbb{E}_{ q(\bx)} \left[ \mathbb{E}_{r(\bz|\bx)} \left[ -\log  q(\bx) - \log q(\bz|\bx) + \log r(\bz |\bx) \right] \right] \nonumber \\
    &= \mathbb{E}_{ q(\bx)} \left[ \mathbb{E}_{r(\bz|\bx)} \left[ -\log  q(\bx) - \log \frac{q(\bx|\bz) q(\bz)}{ q(\bx)} + \log r(\bz |\bx) \right] \right] \nonumber \\
    &= \mathbb{E}_{ q(\bx)} \left[ \mathbb{E}_{r(\bz|\bx)} \left[ - \log q(\bx|\bz) -\log q(\bz) + \log r(\bz |\bx) \right] \right]. 
    \label{eqn:ch5_appendix_entropy_upperbound}
\end{align}
The bound in Equation \eqref{eqn:ch5_appendix_entropy_upperbound} is tractable if sampling from $q(\bx)$ and $r(\bz|\bx)$ is feasible. Both bounds (Eqs. \ref{eqn:ch5_appendix_entropy_upperbound} and \ref{eqn:ch5_methods_entropy_lowerbound_1}) show similarities to the derivation of EP \cite{minka2001} and variational Bayesian inference \cite{jordan1999_intro_vi_graphical_models}. However, note that the lower bound in Equation \eqref{eqn:ch5_methods_entropy_lowerbound} is connected to the objective in EP, although EP only minimizes $D_{\text{KL}}(q \| r)$ locally.
The bound derived with $q(\bx)$ results in a tighter bound compared with variational autoencoders with $q(\bx|\bz)$, as $\mathbb{H}[q(\bx)] \geq \mathbb{H}[q(\bx|\bz)]$ (for details, see \cite{cover2006}).

\section{Estimating the relative increase of the KL divergence}
\label{sec:ch5_appendix_si_est_kl_increase}

The relative increase of the KL divergence induced by decreasing the temperature is denoted as in Equation \eqref{eqn:ch5_methods_rel_kl_inc_ext}, with
\be
c_k = \frac{\log(Z(\beta_{k+1})) - \log(Z(\beta_{k})) + (\beta_{k+1} - \beta_{k}) \left\langle U(\bx) \right\rangle_{q(\bx,\bz)}}{\log Z(\beta_k) + \beta_k\left\langle U(\bx) \right\rangle_{q(\bx,\bz)} - \left\langle \log r(\bz|\bx) \right\rangle_{q(\bx,\bz)} - \mathbb{H}(q(\bx,\bz))}.
\nonumber
\ee

The following addresses the estimation of $\log(Z(\beta_{k+1})) - \log(Z(\beta_{k}))$ with $\Delta \beta_k = \beta_{k+1} - \beta_k$:
\begin{align}
    Z(\beta_k + \Delta \beta_k) &= \int e^{-(\beta_k + \Delta \beta_k) U(\bx)} d\bx \\ \nonumber
    &= \int \frac{e^{-(\beta_k + \Delta \beta_k) U(\bx)}}{ \frac{e^{-\beta_k U(\bx)}}{Z(\beta_k)}}  \frac{e^{-\beta_k U(\bx)}}{Z(\beta_k)} ~d\bx \\ \nonumber
    &= Z(\beta_k) \int e^{-\Delta \beta_k U(\bx)} p\indtarg(\bx;\beta_k) ~d\bx \\ \nonumber
    &= Z(\beta_k) \int e^{-\Delta \beta_k U(\bx)} \frac{ e^{-\beta_k U(\bx)} ~ r(\bz|\bx)}{q(\bx|\bz) ~ q(\bz)} q(\bx,\bz) ~d\bx ~d\bz. \\ \nonumber
\end{align}
We are interested in $\log(Z(\beta_{i+k})) - \log(Z(\beta_{k}))$. Therefore, we write:
\begin{align}
\label{eqn:ch5_appendix_log_z_diff_importancesampling}
    \log(Z(\beta_{i+1})) - \log(Z(\beta_{i})) &= \log \int e^{-\Delta \beta U(\bx)} \underbrace{ \frac{ e^{-\beta_i U(\bx)} ~ r(\bz|\bx)}{q(\bx|\bz) ~ q(\bz)}}_{w} q(\bx,\bz) ~d\bx ~d\bz \\ \nonumber
    &\approx \log \sum_{i=1}^N e^{-\Delta \beta U(\bxi)} W^{(i)}.
\end{align}
Equation (\ref{eqn:ch5_appendix_log_z_diff_importancesampling}) depicts a noisy Monte Carlo estimator for $\log(Z(\beta_{k+1})) - \log(Z(\beta_{k}))$ based on importance sampling \cite{kahn1951importancesampling} with the following normalized weights:
\begin{equation}
    \label{eqn:ch5_appendix_w_unnormalized}
    W^{(i)} = \frac{w^{(i)}}{\sum w^{(i)}} \quad\text{with}\quad w^{(i)} \propto \frac{ e^{-\beta_i U(\bx)} ~ r(\bz|\bx)}{q(\bx|\bz) ~ q(\bz)}.
\end{equation}
As $e^{-\beta_i U(\bx)} r(\bz|\bx)$ may be small for samples $(\bxi,\bzi) \sim q(\bx,\bz)$, we use instead $\log \bar w^{(i)}$ with $\log \bar w^{(i)} = \log w^{(i)} - a$ and $a = \max \{ \log w^{(i)} \}$ to avoid numerical issues.

Whereas above we showed an approximate estimator for $\log(Z(\beta_{k+1})) - \log(Z(\beta_{k}))$, the following addresses $\log(Z(\beta_{k}))$.
To estimate the relative increase in the KL divergence, one requires the normalization constant as mentioned in Equation \eqref{eqn:ch5_methods_rel_kl_inc_ext}. Multistage sampling \cite{valleau1972} provides a way to approximate $Z(\beta_i)$, given all previous $Z(\beta_{k})$ with $k<i$ and $\beta_{i} > \beta_{i-1}$:
\be
\frac{Z(\beta_i)}{Z(0)} =  \frac{Z(\beta_1)}{Z(\beta_0)} \cdot \frac{Z(\beta_2)}{Z(\beta_1)} \cdots \frac{Z(\beta_{i-1})}{Z(\beta_{i-2})}.
\ee
The ratios $\frac{Z(\beta_{k-1})}{Z(\beta_{k-2})}$ are given by  Equation \eqref{eqn:ch5_appendix_log_z_diff_importancesampling}.
The remaining component to be estimated is $Z(0)$, as we utilize the expression from Equation \eqref{eqn:ch5_appendix_log_z_diff_importancesampling} to estimate the ratios of normalization factors. To avoid learning parametrizations $\btheta$ yielding almost uniform $q(\bx)$ on an infinite domain, which occurs in the limit when $\beta =0$, we define the following auxiliary potential to restrict the domain:
\begin{align}
    U_{aux}(\bx) = 
\begin{cases}
U(\bx),& \text{if } \bx \in [-b, b]^{\dim(\bx)}\\
-\frac{u}{\beta} \bx,              & \bx < -b \\
\frac{u}{\beta} \bx,              & \bx > b,
\end{cases}
\end{align}
with $u = \num{10e2}$. The above extension does not influence the potential energy $U(\bx)$ at relevant temperatures.

The initial $Z(\beta_0)$ is computed with importance sampling. This is done only once upon convergence of $(\btheta,\bphi)$ for $\beta_0$:
\begin{align}
    Z(\beta_0) &= \int e^{-\beta_0 U(\bx)} ~d\bx \\ \nonumber
    &= \int  e^{-\beta_0 U(\bx)} ~r(\bz |\bx) ~d\bx ~d\bz \\ \nonumber
    &= \int \underbrace{\frac{e^{-\beta_0 U(\bx)}~r(\bz |\bx)}{q(\bx,\bz)}}_{w}~q(\bx,\bz) ~d\bx ~d\bz.
\end{align}
With samples $(\bxi,\bzi) \sim q(\bx,\bz)$, we obtain the following unnormalized weights:
\be
w^{(i)} = \frac{e^{-\beta_0 U(\bxi)}~r(\bzi |\bxi)}{q(\bxi,\bzi)},
\ee
or $\log w^{(i)} = -\beta_0 U(\bxi) + \log r(\bzi |\bxi) - \log q(\bxi,\bzi)$. Then,
\be
\log Z(\beta_0) = -\log N + \log \sum_{i=1}^N e^{\log w^{(i)} - c } +c,
\ee
with $c = \max (\log w^{(i)})$.

\section{ALA-2 coordinate representation}
\label{sec:ch5_appendix_ala2_coordinate_representation}

\begin{figure}
    \centering
    \includegraphics[width=\textwidth]{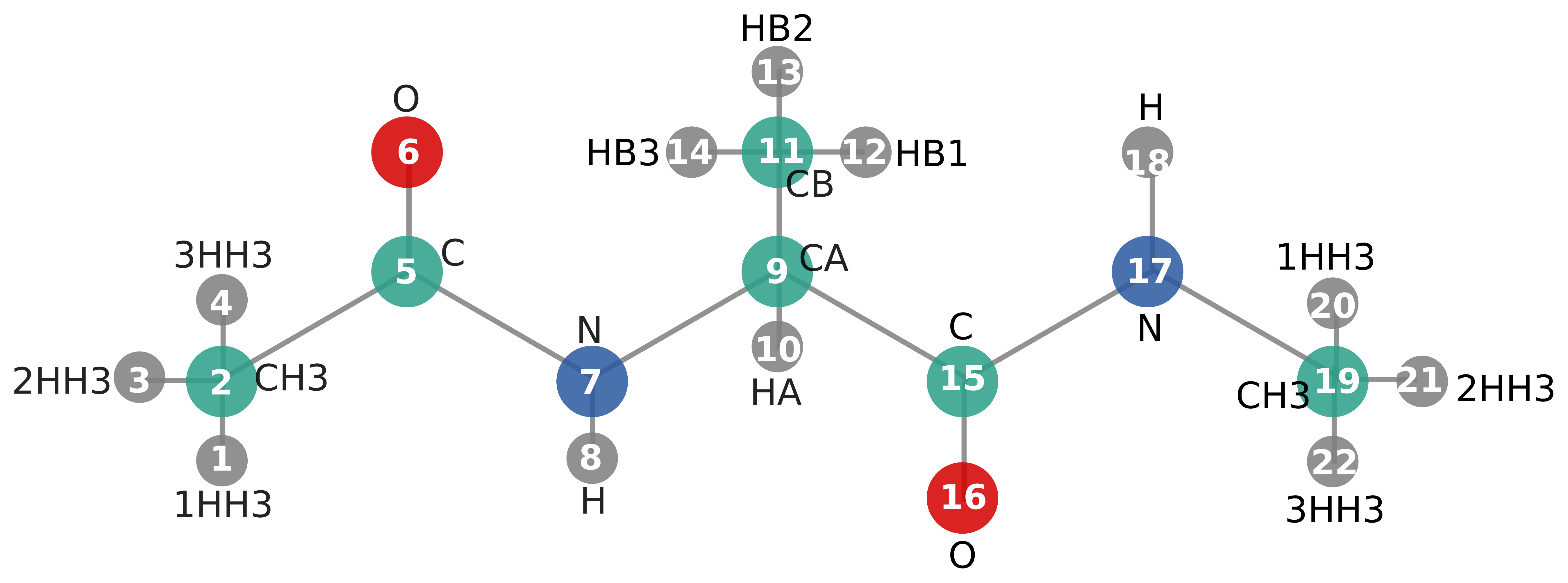}
    \caption{ALA-2 structure with numbered atoms as used for decomposing $\bx$.}
    \label{fig:ch5_appendix_ala_2_structure}
\end{figure}

We show the structure of the ALA-2 petpide in  \figref{fig:ch5_appendix_ala_2_structure}. The numbers in the circles, which depict the involved atoms of ALA-2, correspond to the order in which we assemble block-wise the Cartesian coordinates $(x_i,y_i,z_i)$ of atom $i$ to
\[
\bx = (x_1, y_1,z_1, x_2,y_2,\dots,x_{22},y_{22},z_{22})^T,
\]
where $i$ is the atom number as depicted in \figref{fig:ch5_appendix_ala_2_structure}. For removing rigid-body motion, we fix the Cartesian coordinates $(x_6,y_6,z_6)$ of atom 6, $(x_9,y_9)$ of atom 9, and $(y_{15})$ of atom 15. The employed PDB structure file is available online at \url{https://github.com/cics-nd/predictive-cvs/blob/master/data_peptide/ala-2/ala2_adopted.pdb}.

\section{Simulation of ALA-2}
\label{sec:ch5_appendix_ala2_sim_details}

The procedure for generating a reference trajectory for computing reference observables of ALA-2 is identical to that employed in \cite{schoeberl2019_pcvs}, whereas the data generation approach relies on \cite{shell2012}.
The utilized interaction force field is AMBER ff96 \cite{sorin2005, depaul2010, allen1989}, resolved by an implicit water model based on the  generalized Born  model \cite{onufriev2004, still1990}. Incorporating an explicit water model would, obviously, provide trajectories that would yield observables closer to the \emph{experimental} reference. 
An Andersen thermostat is used to maintain fluctuations around the desired temperature $T=\SI{330}{K}$. All reference simulations are carried out using Gromacs~\cite{berendsen1995, lindahl2001, spoel2005, hess2008, sander2013, prall2015, abraham2015}. The time step is $\Delta t = \SI{1}{fs}$, with a preceding equilibration phase of $\SI{50}{ns}$. Thereafter, a trajectory snapshot is taken  every $\SI{10}{ps}$.  Rigid-body motions have been removed from the Cartesian coordinates.

\section{Observable estimation for ALA-2}
\label{sec:ch5_appendix_ala2_observable_estimation}

We are interested in estimating observables based on predictive models, in contrast to those obtained through reference MD simulations. In general, observables are evaluated as ensemble (MC) or phase (MD) averages, $\int a(\bx) p\indtarg(\bx) ~d \bx$, by making use of $q\indtheta(\bx)$ and samples drawn by ancestral sampling.
We illustrate the radius of gyration (Rg) \cite{fluitt2015_force_field_comparison, shell2012}, given as:
\be
a_{\text{Rg}}(\bx) = \sqrt{\frac{\sum_p m_p \| \bx_p - \bx_{\text{COM}} \|^2}{\sum_p m_p}}.
\label{eqn:ch5_appendix_observable_rg}
\ee
The sum in Equation \eqref{eqn:ch5_appendix_observable_rg} considers all system atoms $p=1,\ldots, P$, with the atom mass $m_p$ and Cartesian coordinate $\bx_p$ of each atom. The center of mass of the peptide is denoted by  $\bx_{\text{COM}}$. A histogram of $a_{\text{Rg}}(\bx)$ reflects the statistics of the peptide's average size, which characterize its various conformations \cite{fluitt2015_force_field_comparison}.

\section{Gradient normalization}
\label{sec:ch5_appendix_method_grad_norm}

During optimization of the objective in the context of atomistic systems, we encounter significant forces, $\mathbf{F}(\bx)$. These differ in magnitude owing to sampling atomistic realizations, which induce, e.g., relatively small distances between bonded atoms. This leads to extreme force components. Gradient normalization \cite{chen2017_gradienta, goodfellow2016_dl} circumvents disruption of the current set of learned parameters $(\btheta, \bphi)$ via a single component attached with extreme magnitudes, owing to, e.g., short bonded distances. Once training proceeds, and predicted atomistic realizations are closer to reasonable ones, the gradient normalization becomes redundant, affecting only gradients in extreme settings where the absolute values of $\mathbf{F}(\bx) \geq \num{1e15}$. After an initial learning phase, such extreme magnitudes do not occur, and thus the normalization does not affect or distort the physics induced by evaluating the force field $\mathbf{F}(\bx)$. 

Given a batch of $I$ samples $\left\{\bxi \right\}_{i=1}^I$ obtained from $q\indtheta(\bx)$, we estimate the gradient of the objective, $\mathbf{g}^{i}(\bxi)$ and calculate its $\ell$-2 norm:
\begin{equation}
l^i = \| \mathbf{g}^{i} \|_2.
\end{equation}
The average gradient norm is $\bar l = 1/I \sum_{i=1}^I l^i$, and we allow a maximal gradient norm based on the mean with $l_{\text{max}} = \kappa \dot \bar l$, $\kappa =3.0$. $\kappa $ was determined by an empirical study. Those gradients with $l^i > l_{\text{max}}$ are normalized such that
\be
\boldsymbol{g}^{i}_{n} = \frac{l_{\text{max}}}{l_i} \mathbf{g}^{i}.
\label{eqn:ch5_appendix_grad_re_scaling}
\ee
As mentioned earlier, realistic atomistic systems at relevant temperatures are not exposed to $\ell$-2 norms of gradients differing more as twice as compared with the gradient with the lowest $\ell$-2 norm. Thus, the gradient normalization is inactive when learning realistic atomistic configurations.